\documentclass[10pt,journal,compsoc]{IEEEtran}
\ifCLASSOPTIONcompsoc
  \usepackage[nocompress]{cite}
\else
  \usepackage{cite}
\fi

\usepackage{url}
\usepackage{graphicx}
\usepackage{amsmath} 
\usepackage{bm}
\usepackage{amsfonts,amssymb} 
\usepackage{booktabs}
\usepackage{bbding}
\usepackage{makecell}
\usepackage{xcolor}
\usepackage{wasysym}
\usepackage{multirow}

\usepackage{subcaption}
\usepackage[most]{tcolorbox}

\usepackage{adjustbox}
\usepackage[edges]{forest}

\usepackage[pagebackref=false,breaklinks=true,colorlinks,bookmarks=false,urlcolor=magenta]{hyperref}
\newcommand{\smallminus}{\includegraphics[height=0.6em]{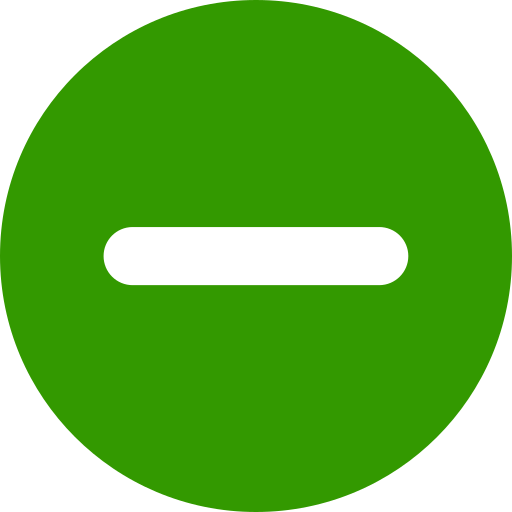}}
\newcommand{\smallsnow}{\includegraphics[height=0.7em]{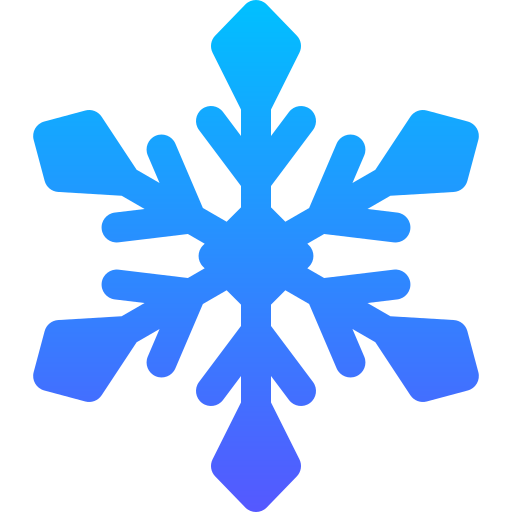}}
\newcommand{\smallfire}{\includegraphics[height=0.7em]{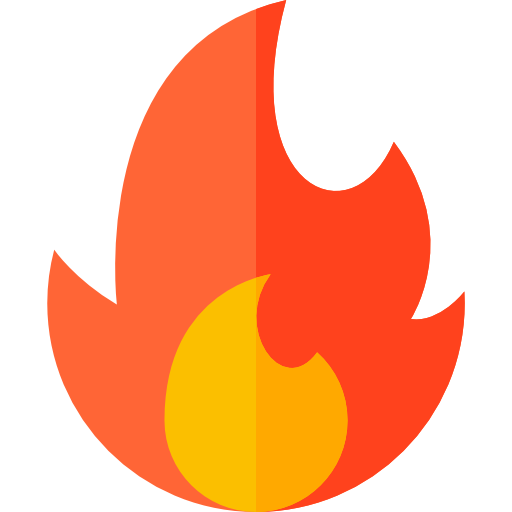}}
\newcommand{\minus}{\includegraphics[height=0.7em]{icon/minus3.png}}
\newcommand{\snow}{\includegraphics[height=0.8em]{icon/snow.png}}
\newcommand{\fire}{\includegraphics[height=0.8em]{icon/fire.png}}
\newcommand{\red}{\textcolor{red}}
\newcommand{\blue}{\textcolor{blue}}

\newcommand{\etal}{\textit{et} \textit{al}.}
\newcommand{\eg}{\textit{e.g.}}
\newcommand{\ie}{\textit{i.e.}}
\newcommand{\myitem}{\noindent\textbf{\emph{$\bullet$}}}

\begin{document}

\title{
Has Multimodal Learning Delivered \\Universal Intelligence in Healthcare?
\\A Comprehensive Survey}

\author{Qika Lin,~\IEEEmembership{Member,~IEEE,}
        Yifan Zhu,~\IEEEmembership{Member,~IEEE,}
        Xin Mei,
        Ling Huang,
        Jingying Ma,\\
        Kai He,~\IEEEmembership{Member,~IEEE,}
        Zhen Peng,
        Erik Cambria,~\IEEEmembership{Fellow,~IEEE,}
        Mengling Feng,~\IEEEmembership{Senior Member,~IEEE}
\IEEEcompsocitemizethanks{
\IEEEcompsocthanksitem It is updating online at: \url{https://github.com/DeepReasoning/aihealth}.
\IEEEcompsocthanksitem Qika Lin, Ling Huang, Jingying Ma, Kai He, and Mengling Feng are
with the Saw Swee Hock School of Public Health, National
University of Singapore, 117549, Singapore.
\IEEEcompsocthanksitem Yifan Zhu is with the School of Computer Science, Beijing University of Posts and Telecommunications, Beijing, 100876, China.
\IEEEcompsocthanksitem Xin Mei is with the School of Automation, Northwestern Polytechnical University, Xi’an, China.
\IEEEcompsocthanksitem Zhen Peng is with the School of Computer Science and Technology, Xi’an Jiaotong University, Xi’an, Shaanxi 710049, China.
}
}



\IEEEtitleabstractindextext{
\begin{abstract}

The rapid development of artificial intelligence has constantly reshaped the field of intelligent healthcare and medicine.
As a vital technology, multimodal learning has increasingly garnered interest due to data complementarity, comprehensive modeling form, and great application potential.
Currently, numerous researchers are dedicating their attention to this field, conducting extensive studies and constructing abundant intelligent systems.
Naturally, an open question arises that \emph{has multimodal learning delivered universal intelligence in healthcare?}
To answer the question, we adopt three unique viewpoints for a holistic analysis.
Firstly, we conduct a comprehensive survey of the current progress of medical multimodal learning from the perspectives of datasets, task-oriented methods, and universal foundation models.
Based on them, we further discuss the proposed question from five issues to explore the real impacts of advanced techniques in healthcare, from data and technologies to performance and ethics.
The answer is that current technologies have \textbf{NOT} achieved universal intelligence and there remains a significant journey to undertake.
Finally, in light of the above reviews and discussions, we point out ten potential directions for exploration towards the goal of universal intelligence in healthcare.

\end{abstract}

}
\maketitle
\IEEEdisplaynontitleabstractindextext
\IEEEpeerreviewmaketitle

\IEEEraisesectionheading{\section{Introduction}}
\IEEEPARstart{R}{ecent} years have seen the remarkable progress of artificial intelligence (AI) across the healthcare and medicine domain~\cite{he2023survey}.
AI techniques have demonstrated substantial potential in various medical scenarios, including medical imaging analysis~\cite{guan2021domain}, disease diagnosis~\cite{nayak2020automated}, drug discovery~\cite{sadybekov2023computational}, personalized treatment~\cite{chi2022producing}, and medical QA (question-answering)~\cite{singhal2023towards}, aiming to provide automated and customized expert-level advice or recommendations to alleviate the burden on both patients and physicians.
Nonetheless, these studies or applications typically consider only single-modality data, \eg, medical image or text, which could result in diminished performance and may not accurately represent authentic application scenarios~\cite{acosta2022multimodal}.

\begin{figure*}[t]
\centering
\includegraphics[width=0.9\linewidth]{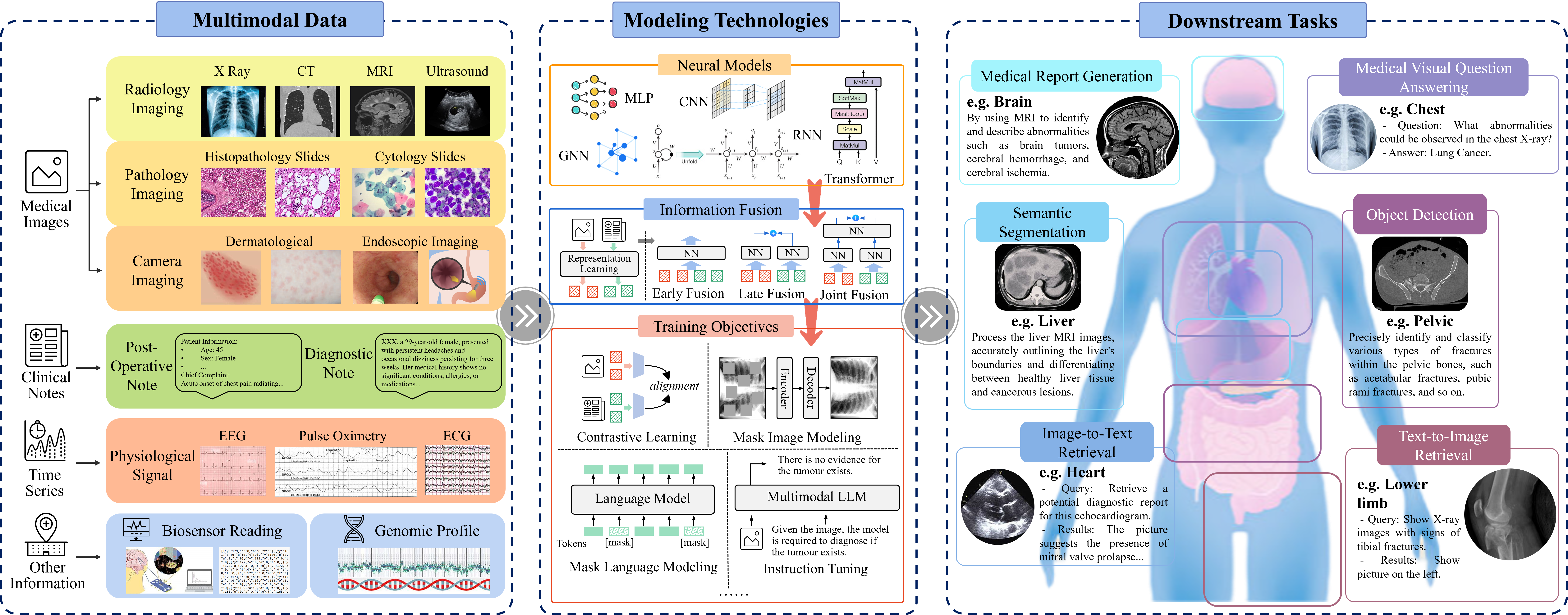}
\caption{Illustration of multimodal learning in healthcare, from perspectives of multimodal data and modeling technologies to downstream tasks.}
\label{fig_intro}
\end{figure*}

As the healthcare domain ceaselessly produces an increasing volume and variety of data, ranging from medical images and clinical notes to genomic profiles and biosensor readings, the need for effective multimodal learning approaches becomes paramount~\cite{shrestha2023medical,pei2024leveraging,messina2022survey,acosta2022multimodal}.
On the one hand, multimodal AI models, capable of integrating and learning from these heterogeneous data streams, hold the promise of unlocking a comprehensive and nuanced understanding of complex medical phenomena.
By capturing complementary semantic information~\cite{wang2018tienet} (as shown in Figure~\ref{fig_intro}) and intricate relationships across different modalities~\cite{DBLP:journals/pami/BaltrusaitisAM19}, these models provide clinicians with a holistic view of patients' conditions, enabling more proactive monitoring, accurate diagnoses, and personalized treatment plans.
On the other hand,
multimodal learning further broadens the application prospects of intelligent models in the healthcare field.
For instance, if a patient needs to ask about their skin condition, articulating it verbally (\eg, using conventional language QA systems) can be challenging.
A visual question-answering (VQA) system becomes incredibly useful, as it can combine intuitive images uploaded by patients to make more accurate and comprehensive diagnoses.
Given the significant research importance and application value of multimodal healthcare, recent years have witnessed an extensive amount of research dedicated to this particular subject, with a clear rising trend.
The advancement in technologies has progressed from utilizing specific models, such as convolutional neural network (CNN)~\cite{resnet}, recurrent neural network (RNN)~\cite{r23}, graph neural network (GNN)~\cite{kipf2016semi}, and Transformer~\cite{transformer}, to the adoption of a strategy involving pre-training and fine-tuning.
The latter has emerged as the prevailing focus and trending topic,
which is inspired by the powerful foundational models (FMs) in the general domain, like CLIP~\cite{radford2021learning}, ChatGPT\footnote{\url{https://openai.com/chatgpt}}, GPT-4\footnote{\url{https://openai.com/gpt-4}}, and multimodal large language model (MLLM)~\cite{DBLP:journals/corr/abs-2306-13549}.
These studies have made significant advancements in numerous tasks of multimodal healthcare, \eg, multi-modality image fusion, report generation (RG), VQA, cross-modal retrieval, text-augmented image processing, and cross-modal image generation.
The evolution has ultimately led to the development of FMs capable of handling various medical tasks.
We summarize the overall architecture in Figure~\ref{fig_intro}.

Despite these seemingly enormous achievements,
it is unclear how far existing research has progressed.
More importantly, the trust among doctors and patients in applying existing methods to real-world scenarios is a significant question~\cite{acosta2022multimodal,hu2024omnimedvqa}.
To this end, we carry out this survey to answer the following open research question: 
\emph{has multimodal learning delivered universal intelligence in healthcare?}, \ie, has multimodal learning delivered an advanced AI healthcare form with a broad range of cognitive abilities, comprehension of various situations, and practical applications?
By answering this question, we aim to provide researchers with a comprehensive global overview of the advancements made towards achieving this objective, the remaining challenges to be addressed, and the necessary steps to be taken.
To achieve this, we conduct our analysis from the following three dimensions:
1) We first comprehensively review the current progress of medical multimodal learning from the perspectives of datasets, task-oriented techniques, and universal FMs.
2) We discuss the open question from five issues to explore the real impacts of current advanced techniques in healthcare applications, from data and technologies to performance and ethics. We find that current technologies have \textbf{NOT} achieved this goal.
3) Drawing upon the above reviews and discussions, we summarize ten prospective directions that present opportunities for deeper investigation into universal intelligence in healthcare.

There are some existing surveys related to the topic of multimodal learning in healthcare.
For example, Shrestha~\etal~\cite{shrestha2023medical} explored the various aspects and advancements of medical vision language pre-training.
Pei~\etal~\cite{pei2024leveraging} surveyed multi-modal learning on biomolecules, which primarily explores the technical advancements in integrating biomolecule sequences, 2D graphs, and 3D structures with natural language processing (NLP) techniques.
Messina~\etal~\cite{messina2022survey} reviewed the RG for medical images and 
Zhao~\etal~\cite{zhao2023clip} made a survey about relevant studies based on CLIP~\cite{radford2021learning} architecture for medical imaging.
Acosta~\etal~\cite{acosta2022multimodal} surveyed multimodal biomedical AI, which mainly focuses on the perspectives of data and applications while there are no technical details involved.
However, it should be noted that these studies only summarize a specific portion of the entire multimodal research studies in healthcare and do not carry out in-depth discussions to explore whether current multimodal learning techniques can realize universal intelligent healthcare.
To the best of our knowledge, this is the first comprehensive and systematic review on AI healthcare multimodal learning,
covering a broad range of topics including datasets, task-oriented methodologies, contrastive FMs, multimodal large language models (MLLMs), as well as discussions on future directions. 
Through this panoramic and in-depth analysis, we conclude that current technology is not yet capable of achieving universal medical AI. We encourage researchers from different domains to collaborate and advance this important journey forward.

The following parts of the paper are organized as follows:
(\S \ref{sec_preliminaries}) gives preliminaries, including modalities, applications, and featured databases.
(\S \ref{sec_study}) shows the details of task-oriented methodologies for medical applications.
(\S \ref{sec_cfm}) and (\S \ref{sec_mllm}) introduce details of two types of multimodal FMs, \ie, contrastive FMs and MLLMs, respectively.
(\S \ref{sec_discussion}) gives the discussions on five sub-questions to answer the proposed research question and (\S \ref{sec_challenge}) further highlights ten future research directions.
Finally, (\S \ref{sec_conclusion}) concludes the study.

\section{Preliminaries}
\label{sec_preliminaries}

\subsection{Medical Modalities}

Academically, modality refers to the way things are expressed or perceived, with every form or source of information being categorized as a modality~\cite{DBLP:journals/pami/BaltrusaitisAM19}.
In the healthcare domain, the term \emph{multimodal} data typically pertains to digital records derived from diverse sources such as various machines, sensors, or experts, often represented in distinct formats.
The medical modalities encompass elements such as medical vision, text, audio, and physiological signals, among others~\cite{acosta2022multimodal}.
As shown in Figure~\ref{fig_intro}, the medical vision modality consists of images obtained by different sensors, which are utilized for viewing the conditions or diseases of different organs or tissues.
Within this scope, three types of images are commonly used, namely radiology, pathology, and camera images.
Radiology is frequently employed to capture images of the human body's internal conditions~\cite{zhang2023knowledge}, primarily encompassing components such as X-ray, computed tomography (CT, 2D/3D), magnetic resonance imaging (MRI), positron emission tomography (PET), and ultrasound.
Pathology is the scientific exploration of disease-induced alterations in cellular and tissue structures, which is conducted through the application of microscopy and supplementary laboratory methodologies~\cite{huang2023visual,lu2024visual}.
Beyond these image types, camera images provide a more direct depiction of the patient's condition and are easier to gather, which is particularly effective and valuable in the detection of skin diseases~\cite{zhou2023skingpt}.
Regarding medical text modalities, they generally encompass domain knowledge and data that are easy for humans to understand, collected from sources such as professional books, diagnostic reports, and literature.
Recently, an increasing number of studies have focused on vision-language learning to provide complementary information and enhance performance~\cite{radford2021learning}.

Other medical modalities, such as audio, physiological signals (including electrocardiogram,~\ie, ECG, and electroencephalogram,~\ie, EEG), and electronic health record (EHR),
also play significant roles in intelligent healthcare.
Nonetheless, relevant studies predominantly concentrate on modeling the intricacies within these individual modalities, overlooking the interaction among multiple modalities.
Thus, our work primarily centers on investigating multimodal studies involving medical images and text, and we discuss more comprehensive multimodal applications in \S\ref{sec_challenge}.

\begin{table}[t]  
\setlength\tabcolsep{4pt}  %
  \centering    
  \caption{Task formalization. ITR: image-to-text retrieval, TIR: text-to-image retrieval.
  $\mathcal{R}$ refers to report and $\mathcal{A}$ refers to answer.
  VQA$_D$/VQA$_G$ denote discriminative/generative setting for VQA. TIP$_{Seg}$ is a segmentation example and $\mathcal{S}$ is the segmentation output.
  $\theta$ is the model parameter. $\mathcal{O}$ is the option set. $\mathcal{Q}$ is the question and $\Gamma$ is the database. $\widetilde{\mathcal{V}}$ is the image of the other modality.
  \textbf{G.} denotes whether the task is discriminative ($\circ$) or generative ($\bullet$).
  }
  \label{tab_tasks}%
  \resizebox{0.5\textwidth}{!}{
  \begin{tabular}{r|cl}
    \toprule
    \textbf{Tasks}  &\textbf{G.}& \textbf{Formalization} \\
    \midrule
    \multirow{2}{*}{RG} &\multirow{2}{*}{$\bullet$}& $\mathcal{R}=[\widehat{r}_1,\cdots,\widehat{r}_n], $\\
    &&$\widehat{r}_i=\mathop{\arg\max}_{w_i \in \mathcal{W}}\prod_{j=1}^i p(w_j|\widehat{r}_1,\cdots,\widehat{r}_{j-1},\mathcal{V};\theta)$\\
    \hline
    VQA$_D$  &$\circ$& $\mathcal{A}=\mathop{\arg\max}_{a_i \in \mathcal{O}}p(a_i|\mathcal{V},\mathcal{Q};\theta)$\\
    \hline
    \multirow{2}{*}{VQA$_G$} &\multirow{2}{*}{$\bullet$}& $\mathcal{A}=[\widehat{a}_1,\cdots,\widehat{a}_n],$\\
    &&$\widehat{a}_i=\mathop{\arg\max}_{w_i \in \mathcal{W}}\prod_{j=1}^i p(w_j|\widehat{a}_1,\cdots,\widehat{a}_{j-1},\mathcal{V},\mathcal{Q};\theta)$\\
    \hline
    ITR  &$\circ$& $\mathcal{T}=\mathop{\arg\max}_{t_i \in \Gamma_{t}}p(t_i|\mathcal{V};\theta)$\\
    \hline
    TIR  &$\circ$& $\mathcal{V}=\mathop{\arg\max}_{v_i \in \Gamma_{v}}p(v_i|\mathcal{T};\theta)$\\
    \hline
    \multirow{2}{*}{TIP$_{Seg}$} &\multirow{2}{*}{$\circ$}& $\mathcal{S}=[\widehat{s}_1,\cdots,\widehat{s}_k],$\\
    && $\widehat{s}_i=\mathop{\arg\max}_{s_i \in \mathcal{O}}p(s_i|\mathcal{V},\mathcal{T};\theta)$\\
    \hline
    CIG  &$\bullet$& $\mathcal{V}=\mathop{\arg\max}_{v_i}p(v_i|\widetilde{\mathcal{V}}/\mathcal{T};\theta)$\\

    \bottomrule
    \end{tabular}
  }
\end{table}

\subsection{Mainstream Applications}
\label{sec_mainapps}

\begin{figure*}[t]
\centering
\includegraphics[width=0.8\linewidth]{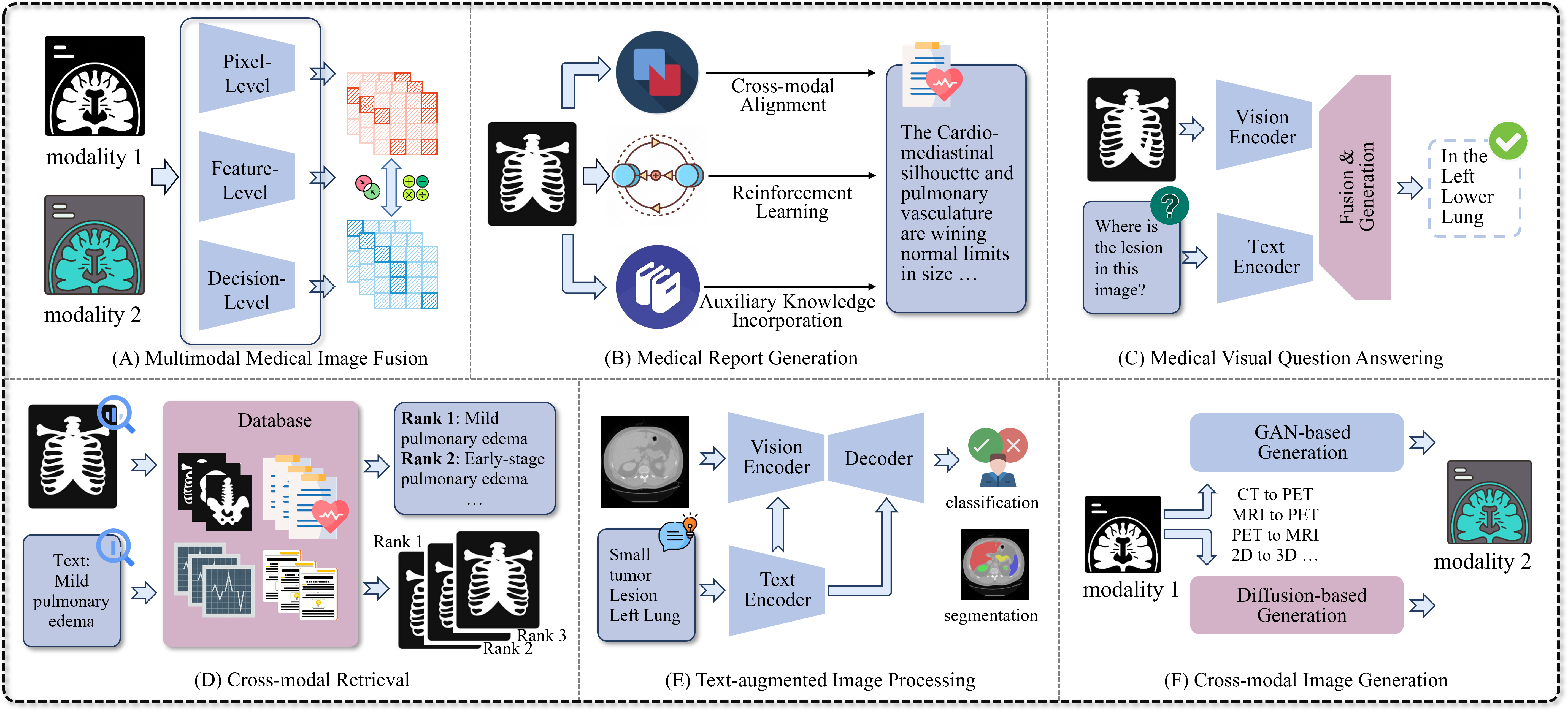}
\caption{Illustration of six mainstream types of task-oriented methodologies for medical applications.}
\label{fig_study_general}
\end{figure*}

Formally, a medical image can be $\mathcal{V}\in \mathbb{R}^{c\times k\times h\times w}$, where $c$, $k$, $h$, $w$ represent the channel, depth, height, and width of images, respectively.
$k=1$ denotes it is a 2D image and $k>1$ indicates it is a 3D image.
Language is represented as $\mathcal{T}=\{w^1,w^2,...,w^m\}$ with max token sequence $m$ and $w_i$ is the word token that from vocabulary $\mathcal{W}$.
There are various multimodal tasks for healthcare using intelligent technologies.
For example, medical image fusion incorporates image features of different modalities.
RG, VQA, cross-modal retrieval (image-to-text and text-to-image), text-augmented image processing (TIP), and cross-modal image generation (CIG) are common tasks in this cross-modal field.
We summarize their formalization in Table~\ref{tab_tasks} and illustrate them in Figure~\ref{fig_study_general}.
Beyond these multimodal tasks, medical multimodal learning can also benefit unimodal tasks, such as medical image classification (IC), semantic segmentation (SS), and object detection (OD).
By integrating data from various sources, medical multimodal learning improves feature representation, enhances contextual understanding, and supports data augmentation, thereby boosting the performance of these unimodal tasks.
We will discuss studies for these applications in \S\ref{sec_cfm}.

\subsection{Featured Databases}
\label{sec_databases}

There are outstanding medical recording databases, which are usually utilized for multimodal healthcare, such as PubMed\footnote{\url{https://pubmed.ncbi.nlm.nih.gov/}}, MIMIC-CXR~\cite{johnson2019mimic}, and UMLS~\cite{bodenreider2004unified}.
PubMed is a free medical literature database, gathering biomedical literature that composes medical journal articles, conference papers, and book chapters.
It provides citations and abstracts, which may include links to the full text.
Similarly, PubMed Central (PMC) provides an archive of full-text articles.
MIMIC-CXR is a comprehensive database comprised of 377K images that correspond to a total of 227K chest radiographic studies.
Each of these studies is accompanied by a detailed radiology report and pertinent chest X-ray (CXR) images.
Authored by radiologists, these reports present a synopsis of their discoveries and typically consist of various sections, including examination, indication, impression, findings, technique, and comparison.
UMLS, short for the unified medical language system, is a comprehensive collection of medical concepts from various lexicon resources.
Each concept is assigned a unique identifier,
which comes with corresponding definitions and numerous synonymous names.
The UMLS also offers insights into the relationships between medical entities with a triplet format, conceptualizing a widespread medical knowledge graph (KG).

Using available resources, certain datasets are curated for specific tasks, such as the prevalent RG and VQA.
We list some representative ones in Table~\ref{tab_rg_datasets} and Table~\ref{tab_vqa_datasets}, which are detailed described in the Appendix and further discussed in \S\ref{sec_discussion}.
They can be utilized for the data construction for MLLMs, as shown in \S\ref{sec_mllm}.
While we only discuss these two kinds of datasets, their utility extends to numerous tasks through the process of transformation. For instance, datasets used for RG can also be utilized for cross-modal retrieval, given their one-to-one correspondence.

\begin{table*}[t!]
  \centering    
  \caption{Representative RG datasets. \# means the number of samples.
  \emph{Col.} means the collection methods, where $\circ$, $\star$, and $\bullet$ represent using synthetical, semi-automatic, and manual manner, respectively.
  }
  \label{tab_rg_datasets}%
  \resizebox{\textwidth}{!}{
  \begin{tabular}{rccccccl}
    \toprule
    \multirow{2}{*}{\textbf{Dataset}}&\multirow{2}{*}{\textbf{Time}}&\multirow{2}{*}{\textbf{\#RG.}}&\multicolumn{2}{c}{\textbf{Images}}&\multirow{2}{*}{\textbf{Source}}&\multirow{2}{*}{\textbf{Col.}}&\multirow{2}{*}{\textbf{Image \& Report Characteristics}}\\
    \cline{4-5}
    &&&\textbf{\#Img.}& \textbf{Modality}&& \\
    \midrule
    IU X-ray~\cite{DBLP:journals/jamia/Demner-FushmanK16}&2016&3.9K&7.4K&CXR&In-House&$\circ$&\emph{Indications}, \emph{findings}, \emph{impression}, \emph{manual encoding}, and \emph{MTI encoding}.\\
    ICLEF-Caption-2017~\cite{DBLP:conf/clef/EickhoffSHM17}&2017&184.6K&184.6K&Multiple&PMC&$\circ$&Image captions in scholarly biomedical articles.\\
    ICLEF-Caption-2018~\cite{DBLP:conf/clef/HerreraEAM18}&2018&232.3K&232.3K&Multiple&PMC&$\circ$&Image captions in scholarly biomedical articles.\\
    PEIR Gross~\cite{DBLP:conf/acl/XingXJ18}&2018&7.4K&7.4K&Pathology&PEIR&$\circ$&Images of gross lesions from sub-categories and one-sentence reports.\\
    ROCO~\cite{DBLP:conf/miccai/PelkaKRNF18}&2018&81.8K&81.8K&Radiology&PMC&$\circ$&Reports are with UMLS CUIs/semantic types for image interrelations.\\
    PadChest~\cite{bustos2020padchest}&2020&109.9K&160.8K&CXR&In-House&$\star$&Reports contain findings, diagnoses, and locations in UMLS taxonomy.\\
    MedICaT~\cite{subramanian2020medicat}&2020&217K&217K&Multiple&PubMed&$\star$&Containing captions, subﬁgures/subcaptions, and inline references.\\
    \multirow{2}{*}{ARCH~\cite{gamper2021multiple}}&\multirow{2}{*}{2021}&\multirow{2}{*}{11.8K}&\multirow{2}{*}{15.1K}&\multirow{2}{*}{Pathology}&PubMed \& &\multirow{2}{*}{$\circ$}& Multiple instance captioning (a caption can relate to multiple images),\\
    &&&&&textbooks&&including diagnostic, detection \& classification, descriptive, etc.\\
    FFA-IR~\cite{DBLP:conf/nips/LiCLWZWCLPLZLSV21}&2021&10.7K&1.0M&FFA&In-House &$\bullet$& Including bilingual (Ch. \& En.) reports and explainable annotations. \\
    CTRG~\cite{tang2024work}&2024&2.8K&8.1K&CT&In-House &$\star$& Reports (template/abnormal contents) of brain and chest CT scans. \\

    \bottomrule
    \end{tabular}
  }
\end{table*}

\begin{table*}[t!]
\centering
\caption{Representative VQA datasets. \# means the number of samples.
\emph{Col.} means the collection methods, where $\circ$, $\star$ and $\bullet$ represent using synthetical, semi-automatic and manual manner, respectively.
\emph{Gen.} denotes if the dataset is generative.
}
\label{tab_vqa_datasets}
\resizebox{\textwidth}{!}{
\begin{tabular}{rcccccccl}
\toprule
\multirow{2}{*}{\textbf{Dataset}}&\multirow{2}{*}{\textbf{Time}}&\multirow{2}{*}{\textbf{\#QA.}}&\multicolumn{2}{c}{\textbf{Images}}&\multirow{2}{*}{\textbf{Source}}&\multirow{2}{*}{\textbf{Col.}}&\multirow{2}{*}{\textbf{Gen.}}&\multirow{2}{*}{\textbf{Characteristics \& Contents}}\\
\cline{4-5}
&&&\textbf{\#Img.}& \textbf{Modality}&&&& \\
\midrule
VQA-Med-2018~\cite{DBLP:conf/clef/HasanLFLML18}&2018&6.4K&2.8K&Radiology&PMC&$\star$&\Checkmark&Rule-based question (location, finding, etc.) generation \& experts check. \\
VQA-RAD~\cite{lau2018dataset}&2018&3.5K&315&Radiology&MedPix&$\bullet$&\Checkmark&About modality, plane, organ system, abnormality, object presence, etc. \\
VQA-Med-2019~\cite{ben2019vqa}&2019&15.2K&4.2K&Radiology&MedPix&$\circ$&\Checkmark&Test set is manually validated; about modality, plane, organ, abnormality.\\
VQA-Med-2020~\cite{DBLP:conf/clef/AbachaDHDM20}&2020&5.0K&5.0K&Radiology&MedPix&$\circ$&\Checkmark&Test set is manually validated; about abnormality.\\
RadVisDial-Silver~\cite{kovaleva2020towards}&2020&455.3K&91.0K&CXR&MIMIC-CXR&$\circ$&\XSolidBrush&Four-choice questions; about 13 abnormalities.\\
RadVisDial-Gold~\cite{kovaleva2020towards}&2020&500&100&CXR&MIMIC-CXR&$\bullet$&\XSolidBrush&Four-choice questions generated by two radiologists.\\
\multirow{2}{*}{PathVQA~\cite{he2020pathvqa}}&\multirow{2}{*}{2020}&\multirow{2}{*}{32.8K}&\multirow{2}{*}{5.0K}&\multirow{2}{*}{Pathology}&Textbooks \&
&\multirow{2}{*}{$\star$}&\multirow{2}{*}{\Checkmark}&First dataset for pathology VQA, using an semi-automated pipeline;\\
&&&&&PEIR&&&about color, location, appearance, shape, etc.\\
VQA-Med-2021~\cite{ben2021overview}&2021&5.5K&5.5K&Radiology&MedPix&$\circ$&\Checkmark&Test set is manually validated; about abnormality.\\
\multirow{2}{*}{SLAKE~\cite{liu2021slake}}&\multirow{2}{*}{2021}&\multirow{2}{*}{14.0K}&\multirow{2}{*}{642}&\multirow{2}{*}{Radiology}&\multirow{2}{*}{3 datasets}&\multirow{2}{*}{$\star$}&\multirow{2}{*}{\XSolidBrush}&Semantically annotated, knowledge-enhanced, and bilingual; about \\
&&&&&&&&plane, modality, position, organ, KG, abnormal, shape, etc.\\
MIMIC-Diff-VQA~\cite{hu2023expert}&2023&700.7K&164.3K&CXR&MIMIC-CXR&$\circ$&\Checkmark&About abnormality, presence, view, location, type, level, and difference.\\
\bottomrule
\end{tabular}
}
\end{table*}

\section{Multimodal Medical Studies}
\label{sec_study}

\subsection{Multi-modality Image Fusion}
Images from a single modality provide limited insight into pathogenetic information within the human body.
A reasonable fusion of multimodal medical images 
significantly contributes to a comprehensive understanding of intricate medical conditions~\cite{zhang2020advances}, allowing clinicians to better delineate anatomical structures, lesions, and abnormalities.
To establish a comprehensive view of how multimodal medical images are combined and analyzed, we introduce existing fusion strategies at the pixel, feature, and decision levels.

\myitem~\textbf{Pixel-level} fusion is a low-level fusion operation that concatenates pixels directly on the original image layer or their corresponding multi-resolution coefficients~\cite{li2017pixel}. These approaches can be classified into three main categories: 1) multi-scale decomposition-based techniques~\cite{shibu2021multi}; 2) sparse representation methods~\cite{liu2015general}, and 3) component substitution techniques~\cite{bhavana2015multi}. However, the outcomes are often impacted by blurring effects that directly affect image contrast~\cite{huang2023deep}. Moreover, there is a high registration requirement for multimodal images and it is usually sensitive to noise, making the pixel-level fusion process time-consuming and challenging.

\myitem~\textbf{Feature-level} fusion is a middle-level fusion strategy, and it is also recognized as the most commonly used strategy in deep learning methods. The common feature-level fusion strategy is to learn a shared representation or a joint embedding space from multiple features, using technologies such as adversarial learning~\cite{safari2023medfusiongan}, co-training~\cite{zhang2023multi}, multi-task learning~\cite{liu2022sf}. More recently, the transformer-based architectures, such as vision Transformer (ViT)~\cite{dosovitskiy2020image}, also show great versatility in handling different types of data, especially for heterogeneous data, and can be leveraged for feature-level fusion tasks by learning a joint representation. While feature-level methods overcome the drawbacks of pixel-based algorithms in terms of contrast, sensitivity to noise, and misregistration, they still have limitations such as spatial distortions~\cite{piella2003general}. For instance, medical images often contain unclear regions due to poor illumination.

\myitem~\textbf{Decision-level} fusion is a high-level fusion strategy. It integrates multiple decisions derived from preliminary classifications and aggregates those decisions finally. Approaches are classified into two main categories: 1) hard fusion methods, which merge logical information membership values, such as model ensembling with majority or average voting~\cite{khan2024hybrid}; and 2) soft fusion methods, where classifiers assign numerical values to reflect their confidence in decisions, or fuzzy classifiers are applied, such as fuzzy voting~\cite{foo2013high}.

\subsection{Medical Report Generation}

Recently, researchers have proposed advanced strategies to enhance the quality and accuracy of automated medical RG,
which can be categorized into three approaches: enhancing cross-modal alignment, improving through reinforcement learning techniques, and integrating auxiliary knowledge.

\myitem~\textbf{Cross-modal Alignment.}
These studies focus on enhancing cross-modal alignment between medical images and reports to improve medical RG.
Najdenkoska~\etal~\cite{vti21} explored learning key topics between images and reports using variational topic inference to enhance semantic coherence.
To facilitate multi-level cross-modal alignments, Li~\etal~\cite{Li_2023_ICCV} unified vision and text modalities into discrete tokens, which are then used to learn global semantic alignment and token-level alignment.
Additionally, contrastive learning techniques~\cite{2022CL1} are also utilized for refined alignment between visual and textual data, enhancing the overall performance.
For example, Wang~\etal~\cite{wang-etal-2023-fine} introduced a phenotype-based contrastive learning framework that learns fine-grained representations, effectively bridging the gap between visual and textual modalities.

\myitem~\textbf{Reinforcement Learning.}
Reinforcement learning (RL) has gained significant traction in text generation due to its ability to use evaluation metrics as rewards, and updates model parameters via policy gradients.
Inspired by this, researchers have applied RL to enhance medical RG models dynamically~\cite{CMAS-RL}, using NLG (natural language generation) metrics-based rewards, \eg, METEOR, ROUGE-L, BLEU, and CIDEr.
To improve factual correctness,
Delbrouck~\etal~\cite{rl-Delbrouc} employed RadGraph~\cite{DBLP:conf/nips/JainASTDBC0LNLR21} to create semantic-based rewards for evaluating the factual accuracy and completeness of reports.
Additionally, Parrer~\etal~\cite{rl-parrer} combined RL with text augmentation to enhance the quality and diversity of radiology reports, using BERTScore~\cite{rl-bertscore} and RadGraph~\cite{DBLP:conf/nips/JainASTDBC0LNLR21} as rewards.
 
\myitem~\textbf{Auxiliary Knowledge Incorporation.}
Recent research has utilized auxiliary signals like medical tags~\cite{CoAtt} and KGs to improve the coherence and accuracy of medical RG.  
For example,
Li~\etal~\cite{li2023dynamic} constructed dynamic KG to integrate both general and specific knowledge and introduced dynamic graph-enhanced contrastive learning to improve visual and textual representations.
Huang~\etal~\cite{huang2023kiut} developed the KiUT model, which integrates clinical knowledge by a symptom graph. It is combined with visual and contextual information through a U-Transformer architecture, significantly enhancing radiology RG.
Hou~\etal~\cite{hou-etal-2023-organ} proposed ORGAN, an observation-guided method using tree reasoning over observation graphs to improve interpretability and coherence.
Additionally, researchers also included patients' prior examination data as anatomically aligned inputs to compare current and previous scans effectively~\cite{bannur2023learning,DBLP:conf/emnlp/HouCXLL23}.

\subsection{Medical VQA}

The study of medical VQA focuses on interpreting medical images alongside spoken-language queries to generate accurate natural-language responses.
It exhibits significant promise for diagnostic assistance and enhancing patient comprehension through educational support.
The key research question revolves around the precise identification and comprehension of important regions within medical images (such as lesions, anomalies, and occupying masses), and the semantic feature space alignment of these regions with the core demands expressed in the textual queries~\cite{OVQA}.

\myitem~\textbf{Knowledge Extraction Frameworks.}
For encoder structures on a single modality, employing learning strategies that allow them to concurrently preserve both pre-existing general visual common sense and domain-specific medical knowledge represents an efficient model-agnostic solution.
The first category employs the meta-learning paradigm, where the VQA model distributes the parameter training process across multiple related tasks to fuse loss gradients,
ultimately adapting the model to medical VQA scenarios, \eg, MAML~\cite{DBLP:conf/miccai/NguyenDNDTT19} for model transfer and MMQ~\cite{DBLP:conf/miccai/DoNTTTN21} for data refinement.
Another approach to integrating medical knowledge with common sense is to perform conditional reasoning, which is designed to learn task-adaptive reasoning skills for different types of VQA tasks.
This reasoning ability can be implemented by a supervised label loss on the embedding fusion of visual and textual encoders~\cite{DBLP:journals/artmed/WangHLQCLR22}.

\myitem~\textbf{Pre-training \& Fine-tuning Frameworks.}
The remarkable advancements in pre-trained language models (LMs), such as BERT~\cite{devlin2019bert}, have also introduced a novel paradigm for medical VQA.
By designing self-supervised learning tasks, LMs can learn from vast text corpora without relying on external annotations, thus significantly reducing the cost of labeling.
When faced with specific downstream VQA tasks, only a small amount of labeled data is required to finetune the model.
In particular, knowledge relationships can also be injected into the pre-training process by non-euclidean encoders such as GNN~\cite{chen2022align}.
Owing to the computational constraints of adjusting a large number of parameters during the fine-tuning phase, recent research has introduced VQA downstream task adaptation strategies that involve freezing parameters of the pre-trained model. 
For example,
Liu~\etal~\cite{liu2023parameter} proposed to establish a VQA adapter that is external to the pre-trained model, thus creating a plug-in for model adaptation to downstream tasks.

\subsection{Cross-modal Retrieval}

The medical cross-modal retrieval studies mainly line in the two categories: 
cross-modal retrieval within images, and retrieval between images and texts.
Cross-modal medical image retrieval involves searching for medical images in a database that have similar visual features to a given query image, thereby facilitating efficient clinical decision-making.
Traditional studies handle the retrieval task by calculating similarities between texture features of different images, such as Radon Transform~\cite{liu2016generating} on X-rays.
Further, Mbilinyi~\etal~\cite{mbilinyi2020cross} suggested the application of deep features for extracting similar medical images from multimodal medical image databases.
The results show that the retrieval performance of deep features obtained by CNNs is superior to the conventional texture features.
Xu~\etal~\cite{xu2022multi} proposed multi-manifold deep discriminative cross-modal hashing for extensive medical image retrieval.
The core aspect is the multi-modal manifold similarity that integrates multiple sub-manifolds based on heterogeneous data, thereby preserving the correlation among instances. 
This approach is both effective and efficient in adaptively retrieving medical images across various modalities.
In general, this class of methods has undergone a transition from traditional feature extraction to efficient retrieval by deep neural networks.

Currently, the cross-modal retrieval between medical images and text, including ITR and TIR, is mainly done by learning embedded representations and calculating similarities by neural networks.
The current major trend is to enhance representations with additional prior domain knowledge, such as the category information~\cite{zhang2022category} and hierarchical semantic associations between disease labels~\cite{ding2023semantic}.

\subsection{Text-augmented Image Processing}

Considering the complementary information contained in the text and image modalities, some studies focus on using text-guided information for medical image processing, including image classification and semantic segmentation.

Some text descriptions about quantity and scale could provide additional supporting information for image understanding and segmentation.
For example, LViT~\cite{DBLP:journals/corr/abs-2206-14718} composes a U-shaped CNN branch and a U-shaped ViT branch for segmentation.
It utilizes medical text annotation to address the limitations in image data quality, guiding the generation of enhanced pseudo labels in semi-supervised learning.
Zhao~\etal~\cite{zhao2024dtan} and Dong~\etal~\cite{dong2024diffusion} introduced text-guided diffusion models for medical image segmentation. These models employ a text-attention mechanism to mitigate the influence of variations in the size and quantity of objects, such as representations of \emph{one}, \emph{many}, \emph{small}, \emph{medium}, and \emph{large}, on the segmentation results.
By concentrating on appropriate textual descriptors, the network is capable of adaptively modulating its focus towards the key characteristics of target objects, ensuring that the segmentation is both accurate and robust to changes in object attributes.
Recently, medical FMs are frequently employed for text-guided image tasks, which is shown in \S\ref{sec_cfm} and \S\ref{sec_mllm}.

\subsection{Cross-modal Image Generation}

In the medical domain, cross-modal image generation (also called modality translation) can be utilized in various scenarios, including education, data augmentation, missing data filling, and pathology research \& understanding.
From the technical perspective, they are categorized into two classes:
GAN-based (generative adversarial network) and diffusion-based.
GAN-based models adopt the principle of adversarial training~\cite{goodfellow2014generative}, involving two networks. The first generator is responsible for creating synthetic instances based on training data. The second discriminator is to differentiate between synthetic and real data.
This competitive dynamic prompts the generator to produce highly realistic samples.
In reality, CT scans emit radiation, potentially causing patient side effects, and their effectiveness in providing detailed images of soft tissue injuries is somewhat restricted.
In contrast, MRI is radiation-free and safer.
Thus, there is growing interest in generating CT images from corresponding MRI ones using GAN models, including perspectives of context-aware~\cite{nie2017medical} and gradient consistency~\cite{hiasa2018cross}.
Also, there are several studies on the generation of other modalities, \eg, CT to PET~\cite{ben2019cross}, MRI to PET~\cite{pan2018synthesizing}, and PET to MRI~\cite{choi2018generation}.

Cross-modal diffusion-based generation models essentially transform the task of direct target generation into predicting random noise at every diffusion step. At its core, the diffusion model contains two critical processes: the forward diffusion process and the reverse denoising process.
The forward diffusion process incrementally introduces Gaussian noise into an instance until it morphs into a sample of random noise. Conversely, the reverse denoising process aims to predict and remove the introduced noise~\cite{ho2020denoising}.
Lyu~\etal~\cite{lyu2022conversion} made full use of denoising diffusion and score-matching strategies based on four different sampling approaches, implementing MRI to CT image synthesis.
The results show that the model generates better synthetic CT images than the CNN and GAN models.
Similarly, Meng~\etal~\cite{meng2022novel} introduced a unified multi-modal conditional score-based generative model to synthesize the missing modality using remaining modalities as conditions.
The model employs only a score-based network to learn different cross-modal conditional distributions and the results show it can more reliably synthesize missing modality images of MRI.

\section{Contrastive Foundation Models (CFMs)}
\label{sec_cfm}

\begin{table*}[t]
\setlength\tabcolsep{2pt}  
\centering
\caption{Representative CFMs. Abbreviations are as follows:
\emph{CPA}: cross-modal prototype alignment,
\emph{DEP}: disease/entity prediction, \emph{ITM}: image-text Matching, \emph{CTR}: cross-lingual text alignment regularization.
``/'' in the \emph{Objectives} \& \emph{Training Process} splits different pre-training stages.
Icons \smallsnow, \smallfire, and \smallminus\; denote the module is frozen, updating, and inexistence when training, respectively.
Their positions correspond image/adapter/language models.
RN is short for ResNet.
}
\label{tab_cfms}
\resizebox{\textwidth}{!}{
\begin{tabular}{rccll}
\toprule
\multirow{2}{*}{\textbf{Model}}&\multirow{2}{*}{\textbf{Time}}&\multirow{2}{*}{\textbf{Modality}}&\textbf{Image/Adapter/Language}&\multirow{2}{*}{\textbf{Objectives (Training Process), Features \& Applications}}\\
&&&\textbf{Model}&\\
\midrule
ConVIRT~\cite{DBLP:conf/mlhc/0004JMML22}&10/2020&Radiology&RN50/--/ClinicalBERT&GCL (\fire\minus\fire); contrastive visual representation from paired descriptive text; IC, zero-shot ITR/TIR.\\
PubMedCLIP~\cite{eslami2023pubmedclip}&12/2021& Multiple&ViT-B-32, RN50/--/BioClinicalBERT&GCL (\fire\minus\fire); fine-tuning CLIP on PubMed articles, followed by VQA-aware model; medical VQA.\\
CheXzero~\cite{tiu2022expert}&09/2022&CXR&ViT-B-32/--/Transformer& GCL(\fire\minus\fire); self-supervised learning on CXR images; (low-resource) IC, \ie, pathology detection.\\
BiomedCLIP~\cite{zhang2023biomedclip}&03/2023&Multiple&ViT-B-16/--/PubMedBERT& GCL (\fire\minus\fire); tuning on very large-scale dataset PMC-15M; (zero-shot) IC, ITR, TIR, VQA.\\
PLIP~\cite{huang2023visual}&03/2023&Pathology&ViT-B-32/--/Transformer& GCL (\fire\minus\fire); pathology data from Twitter; (zero-shot) IC, TIR, ITR, image representations.\\
PathCLIP~\cite{sun2024pathasst}&05/2023&Pathology&ViT-B-16/--/Transformer& GCL (\fire\minus\fire); tuning on PathCap with 207K high-quality image-text pairs; zero-shot IC, TIR.\\
CT-CLIP~\cite{hamamci2024foundation}&03/2024&3D CT&CT-ViT/--/CXR-BERT& GCL (\fire\minus\fire); 3D image abilities; (zero-shot) IC, volume-to-volume/report-to-volume retrieval.\\
PairAug~\cite{xie2024pairaug}&04/2024&CXR&ViT-B-32/–/Transformer& GCL (\fire\minus\fire); inter \& intra data augmentation by ChatGPT and cross-attention maps; (zero-shot) IC. \\

\midrule
GLoRIA~\cite{DBLP:conf/iccv/HuangSLY21}&10/2021&CXR&RN50/--/BioClinicalBERT& GCL+LCL (\fire\minus\fire); text-attention local image representations; zero-shot ITR/IC, low-resource SS.\\
BioViL~\cite{boecking2022making}&04/2022&CXR&RN50/--/CXR-BERT&GCL+MLM (\fire\minus\minus/\fire\minus\fire); language \& vision-language tuning; (zero-shot) IC, SS, phrase grounding.\\
MedCLIP~\cite{DBLP:conf/emnlp/0008WA022}&10/2022&CXR&ViT/--/BioClinicalBERT& \emph{soft} GCL (\fire\minus\fire); soft semantic matching for the false negative issue; (zero-shot) IC, ITR.\\
MGCA~\cite{DBLP:conf/nips/WangZWVY22}&10/2022&CXR&ViT-B-16/--/BioClinicalBERT& GCL+LCL+CPA (\fire\minus\fire); pathological region-level, instance-level \& disease-level CL; IC, SS, OD.\\
\multirow{2}{*}{BioViL-T~\cite{bannur2023learning}}&\multirow{2}{*}{01/2023}&\multirow{2}{*}{CXR}&\multirow{2}{*}{CNN–Transformer/--/CXR-BERT}& GCL+LCL+MLM (\fire\minus\fire); multi-granularity and temporal connectivity modeling of images; \\
&&&& phrase grounding, (zero-shot, temporal) IC, RG, temporal sentence similarity.\\
MedKLIP~\cite{DBLP:conf/iccv/WuZZWX23}&01/2023&CXR&RN50/Transformer/ClinicalBERT & GCL+DEP (\fire\fire\snow); incorporate entities \& their descriptions; (zero-shot) IC, SS, region grounding.\\
KAD~\cite{zhang2023knowledge}&02/2023&CXR&RN50/Transformer/PubMedBERT&  GCL/GCL+DEP (\minus\minus\fire/\fire\fire\snow); medical KG-enhanced; (zero-shot) IC, \ie, disease prediction.\\
PTUnifier~\cite{chen2023towards}&02/2023&Radiology&CLIP-ViT-B/Transformer/RoBERTa& GCL+MLM+ITM (\fire\fire\fire); soft prompts to unify early-fusion and later-fusion; ITR, TIR, VQA, etc.\\
\multirow{2}{*}{Med-UniC~\cite{wan2024med}}&\multirow{2}{*}{05/2023}&\multirow{2}{*}{CXR}&RN50, ViT-B-16(L-32)/MLP/& GCL+CTR (\fire\fire\fire); cross-lingual text alignment regularization for unifying
cross-lingual (English \\
&&&CXR-BERT&\& Spanish) medical vision-language pre-training; (zero-shot) IC, SS, OD.\\
MCR~\cite{wei2023masked}&12/2023&CXR&ViT-B-16/--/BioClinicalBERT& GCL+MLM+MIM (\fire\minus\fire); masked image and text as inputs, additional MLM and MIM; ITR, TIR.\\
MLIP~\cite{li2024mlip}&02/2024&CXR&RN-50, ViT-B-16/--/& GCL+LCL+CPA (\fire\minus\fire); multi-granularity, KG-based LCL; (zero-shot) IC, SS, OD.\\
MAVL~\cite{phan2024decomposing}&03/2024&CXR&RN50/Transformer/ClinicalBERT& GCL+DEP (\fire\fire\snow); disease entities \& their descriptions; (zero-shot) IC, SS, visual grounding.\\
\multirow{2}{*}{KEP~\cite{zhou2024knowledge}}&\multirow{2}{*}{04/2024}&\multirow{2}{*}{Pathology}&\multirow{2}{*}{ViT-B-32(B-16)/--/PubMedBERT}& AdaSP/GCL (\snow\minus\fire/\fire\minus\fire); incorporate a knowledge tree of disease attributes; ITR, TIR, disease \\
&&&& retrieval, zero-shot classification on pathology patches, and zero-shot tumor subtyping on WSIs. \\
DeViDe~\cite{luo2024devide}&04/2024&CXR&ViT-B/LP/Med-KEBERT& GCL (\fire\fire\fire); entity-aware descriptions, global and local CL; (zero-shot) IC, IC, SS.\\
\bottomrule
\end{tabular}
}
\end{table*}

\begin{figure}[t]
  \centering
  \begin{minipage}[t]{0.9\linewidth}
    \large
    \centering
    \includegraphics[width=0.9\linewidth]{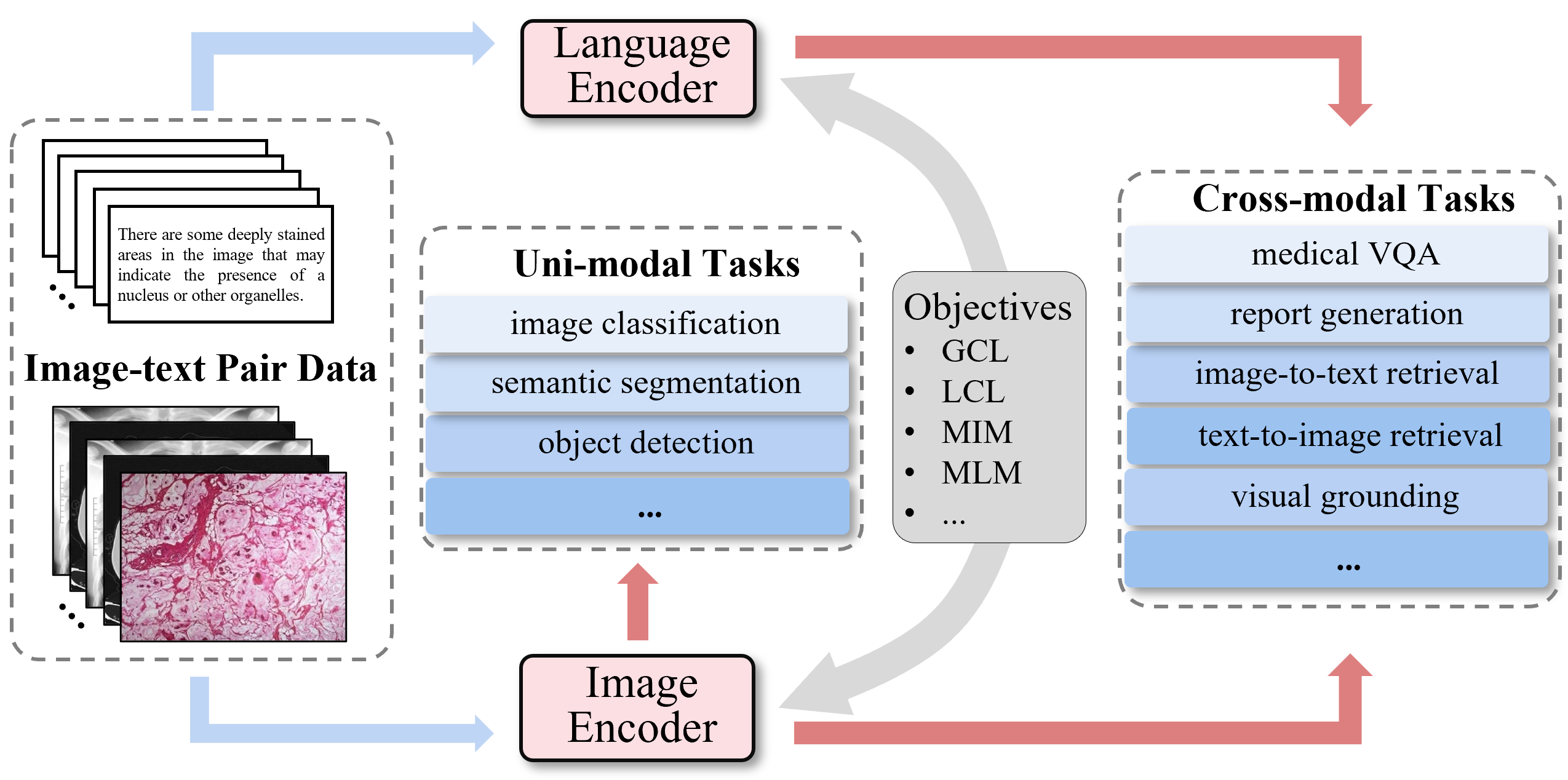}
    \subcaption{The general architecture of contrastive FMs.}
  \end{minipage}
  \begin{minipage}[t]{0.9\linewidth}
    \large
    \centering
    \includegraphics[width=\linewidth]{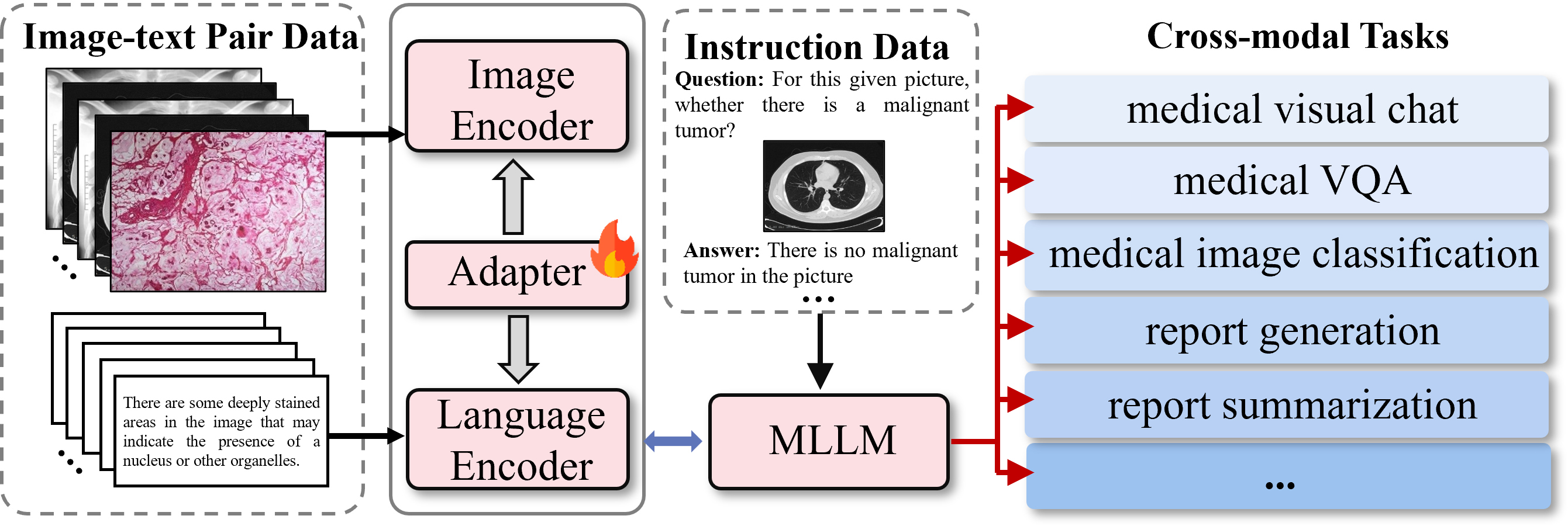}
    \subcaption{The general architecture of MLLMs.}
  \end{minipage}
  \caption{Illustration of FMs in the medical domain.
  }
  \label{fig_fms}
\end{figure}

Given the intrinsic rarity, specificity, and specialized nature of data in the medical field,
it is challenging and unrealistic to take large-scale high-quality annotated data for training.
Therefore, some self-supervised strategies are introduced for building universal FMs~\cite{moor2023foundation}.
FMs typically denote models that acquire the broad representation of general knowledge by undergoing pre-training on large-scale data through self-supervised learning.
Subsequently, they can be refined through fine-tuning.
Figure~\ref{fig_fms} illustrates the general architecture and applications of FMs in the medical domain.
FMs have several key features: 1) pre-training on large generic datasets; 2) self-supervised learning strategies, such as contrastive learning and mask language modeling; 3) universal knowledge representation, meaning FMs learn a generic, task-independent knowledge representation that can be applied to a variety of different downstream tasks with a small amount of fine-tuning.
According to the training strategies and applications, they can be categorized into two types:
contrastive FMs (CFMs) and MLLMs.
CFMs focus on learning a common cross-modal representation space by jointly optimizing the image encoder and text encoder to maximize the similarity score of the positive sample (image-text pair) and minimize the similarity score of the negative sample~\cite{zhao2023clip}.
However, MLLMs focus more on modeling the intrinsic cross-modal relationships, implementing cross-modal computation, and are capable of generating text outputs~\cite{DBLP:journals/corr/abs-2306-13549}.
In this section, we will introduce CFMs and MLLMs will be detailedly elaborated in \S\ref{sec_mllm}.

\subsection{Overview of CFMs}
\label{sec_cfms}

Borrowing the idea of self-supervised contrastive learning (CL) that achieved great success in the computer vision field,
CLIP~\cite{radford2021learning} aligns vision and language semantic representations by pre-training on large-scale image-text pairs, which has greatly promoted the development of visual semantic understanding.
It has sparked interest in its potential applications in the medical field.
The general idea is to use an image encoder (\eg, pre-trained ResNet~\cite{he2016deep} or ViT~\cite{dosovitskiy2020image}) and a language encoder (\eg, RoBERTa~\cite{liu2019roberta}, CXR-BERT~\cite{boecking2022making}, ClinicalBERT~\cite{alsentzer2019publicly}, or PubMedBERT~\cite{gu2021domain}) to obtain their corresponding representations.
Rarely, an adapter serves as the bridge between them.
Subsequently, they are updated using semantic contrast.
We classify these studies into two categories: \emph{CLIP-based} and \emph{CLIP-variant} pre-training.
The former generally uses typical global CL loss for modeling as like the original CLIP, while the latter introduces new optimizing objectives for additional specific concerns.
Some representative medical CFMs are listed in Table~\ref{tab_cfms} and some representative datasets for alignment are in Table~\ref{tab_mllm_dataset}.

\subsection{CLIP-based Pre-training}

Formally, given the image set $\{\mathcal{V}_1,\mathcal{V}_2,\cdots,\mathcal{V}_N\}$ and text set $\{\mathcal{T}_1,\mathcal{T}_2,\cdots,\mathcal{T}_N\}$ where $\mathcal{V}_i$ and $\mathcal{T}_i$ form a image-text pair,
their representations $\mathbf{v}\in \mathbb{R}^{N\times n \times d}$ and $\mathbf{t}\in \mathbb{R}^{N\times m \times d}$ can be obtained by image and text encoders, respectively.
This formalization is similar to \S\ref{sec_mainapps}.
Further, their global representation can be obtained by pooling operation or taken a representative one as $\mathbf{v}^g\in \mathbb{R}^{N \times d}$ and $\mathbf{t}^g\in \mathbb{R}^{N \times d}$.
$\mathcal{D}$, $\mathcal{N}$, and $\mathcal{M}$ are the index sets for the samples, image patches, and text tokens.
Based on them, 
global GL (GCL) employs a comprehensive and holistic perspective for semantic relationships.
Its loss is to directly model the whole semantics between image and text using InfoNCE loss:
\begin{equation}
\mathcal{L}_{\rm GCL} = \mathop{\mathbb{E}}_{i\in \mathcal{D}}\Big[-\log\frac{\exp\big(s(\mathbf{v}_i^g,\mathbf{t}_i^g)\big)/\tau}{\sum_{j=1}^N\exp
\big(s(\mathbf{v}_i^g,\mathbf{t}_j^g)\big)/\tau} \Big].
\end{equation}
$s$ denotes the cosine similarity function and $\tau$ is a pre-set temperature parameter.
Note that it may $s(a,b)\neq s(b,a)$, so $s(\mathbf{t}_i^g,\mathbf{v}_i^g)$ may also be calculated for comprehensive modeling.
Using GCL, many studies learn semantic representations for medical images and their corresponding description texts.
For example, studies on CXR (CheXzero~\cite{tiu2022expert} \& PairAug~\cite{xie2024pairaug}), pathology (PLIP~\cite{huang2023visual} \& PathCLIP~\cite{sun2024pathasst}), radiology (ConVIRT~\cite{DBLP:conf/mlhc/0004JMML22}), and multiple modalities (PubMedCLIP~\cite{eslami2023pubmedclip} \& BiomedCLIP~\cite{zhang2023biomedclip}).
Unlike these approaches concentrating on 2D images,
Hamamci~\etal~\cite{hamamci2024foundation} proposed the first 3D medical imaging dataset CT-RATE with textual reports.
Based on it, CT-CLIP is pre-trained using a 3D encoder for chest CT volume representations and it is then aligned with CXR-BERT outputs.

\subsection{CLIP-variant Pre-training}
Beyond GCL, there are studies with other modeling objectives, which can be summarized as following four types.

\myitem~\textbf{Intra-modal modeling.}
Beyond inter-modal modeling, there are additional objectives focused on intra-modal modeling, where two standard objectives are typically employed.
MIM (mask image modeling)~\cite{wei2023masked} usually uses the mean squared error function to compute the normalized pixel-wise difference between the original image patches $P_i$ and reconstructed image patches $\widetilde{P}_i$. MLM (mask language modeling)~\cite{chen2023towards,bannur2023learning,wei2023masked} are to predict masked tokens based on other given tokens.
They are calculated as follows:
\begin{equation}
\small
\mathcal{L}_{\rm MIM} =\mathop{\mathbb{E}}_{i\in \mathcal{D}}\big[(P_i-\widetilde{P}_i)^2\big],\; \mathcal{L}_{\rm MLM} =\mathop{\mathbb{E}}_{i\in \mathcal{D},j\in \mathcal{M}}\big[ \mathbb{I}(\mathcal{T}_i^j)\cdot f(\mathcal{T}_i^j)\big].
\end{equation}
$\mathbb{I}$ is to indicate where token $T_i^j$ is masked, where $\mathbb{I}(\mathcal{T}_i^j)=1$ if masked.
Otherwise, the value is 0.
$f$ is the loss for the predicted token compared to the original one.
These two approaches acquire internal semantics of modality through reconstruction, enhancing single-modal model capability.

\myitem~\textbf{Multi-granularity.}
Focusing on the global information of images and texts may not be sufficient for capturing fine-grained information.
So based on GCL, its variant of multi-granularity is introduced, where local CL (LCL) is the representative.
It utilizes the global (or local) representations of one modality to align local representations of another.
Local token representations of text and image are $\mathbf{t}_i^j$ and $\mathbf{v}_i^j$, which means the token $j$ of sample $i$.
The corresponding global representation can be $\mathbf{\widetilde{t}}_i^g$ and $\mathbf{\widetilde{v}}_i^g$, obtained by pooling strategy or linear transformation based on $\mathbf{t}_i^g$ and $\mathbf{v}_i^g$.
There are two main LCL approaches (global-to-local as example):
\begin{equation}
\small
\mathcal{L}_{\rm LCL (I2T)} = \mathop{\mathbb{E}}_{i\in \mathcal{D},j\in \mathcal{M}}\Big[-
\log\frac{\exp\big(\sigma(\mathbf{\widetilde{v}}_i^g,\mathbf{t}_i^j)\big)/\tau}{\sum_{k=1}^N\exp
\big(\sigma(\mathbf{\widetilde{v}}_k^g,\mathbf{t}_i^j)\big)/\tau} \Big],
\end{equation}
\begin{equation}
\small
\mathcal{L}_{\rm LCL (T2I)} = \mathop{\mathbb{E}}_{i\in \mathcal{D},j\in \mathcal{N}}\Big[-
\log\frac{\exp\big(\sigma(\mathbf{\widetilde{t}}_i^g,\mathbf{v}_i^j)\big)/\tau}{\sum_{k=1}^N\exp
\big(\sigma(\mathbf{\widetilde{t}}_k^g,\mathbf{v}_i^j)\big)/\tau} \Big].
\end{equation}
The former uses global image information as the anchor, aligning it with corresponding local text counterparts.
MGCA~\cite{DBLP:conf/nips/WangZWVY22} and BioViL-T~\cite{bannur2023learning} follow the latter manner, viewing the global text representation as the anchor.
GLoRIA~\cite{DBLP:conf/iccv/HuangSLY21} uses both manners, where local representations are obtained by text token-based attentive image pooling.

In a different manner,
MLIP~\cite{li2024mlip} introduces local token-knowledge-patch alignment using a medical KG, \ie, UMLS.
The cross-attention is used to match knowledge and image or text, with the input of pre-trained entity embeddings of UMLS and feature of image patch or text token.
Processed representations are compared with the original ones for knowledge alignment, from both image and text sides.
Thus, MLIP is in a manner of local-local contrast rather than other global-local methods.
Beyond LCL, MGCA~\cite{DBLP:conf/nips/WangZWVY22} also introduces cross-modal prototype alignment, which assumes that images with the same disease would have similar disease-level representations.
It has a higher level than the global instance and the local tokens.
To realize it, MGCA pre-defines trainable embeddings for cross-modal prototypes with a certain number, which can be used for \emph{pseudo-label} calculation with global text and image representations.
Finally, the model is optimized by aligning the text and image pseudo-labels.
Similarly, MLIP also introduces a higher level, \ie, category-level,
where KG embeddings and original features are guided for category clustering.

\begin{figure}[t]
  \centering
  \begin{minipage}[t]{0.49\linewidth}
    \large
    \centering
    \includegraphics[width=0.9\linewidth]{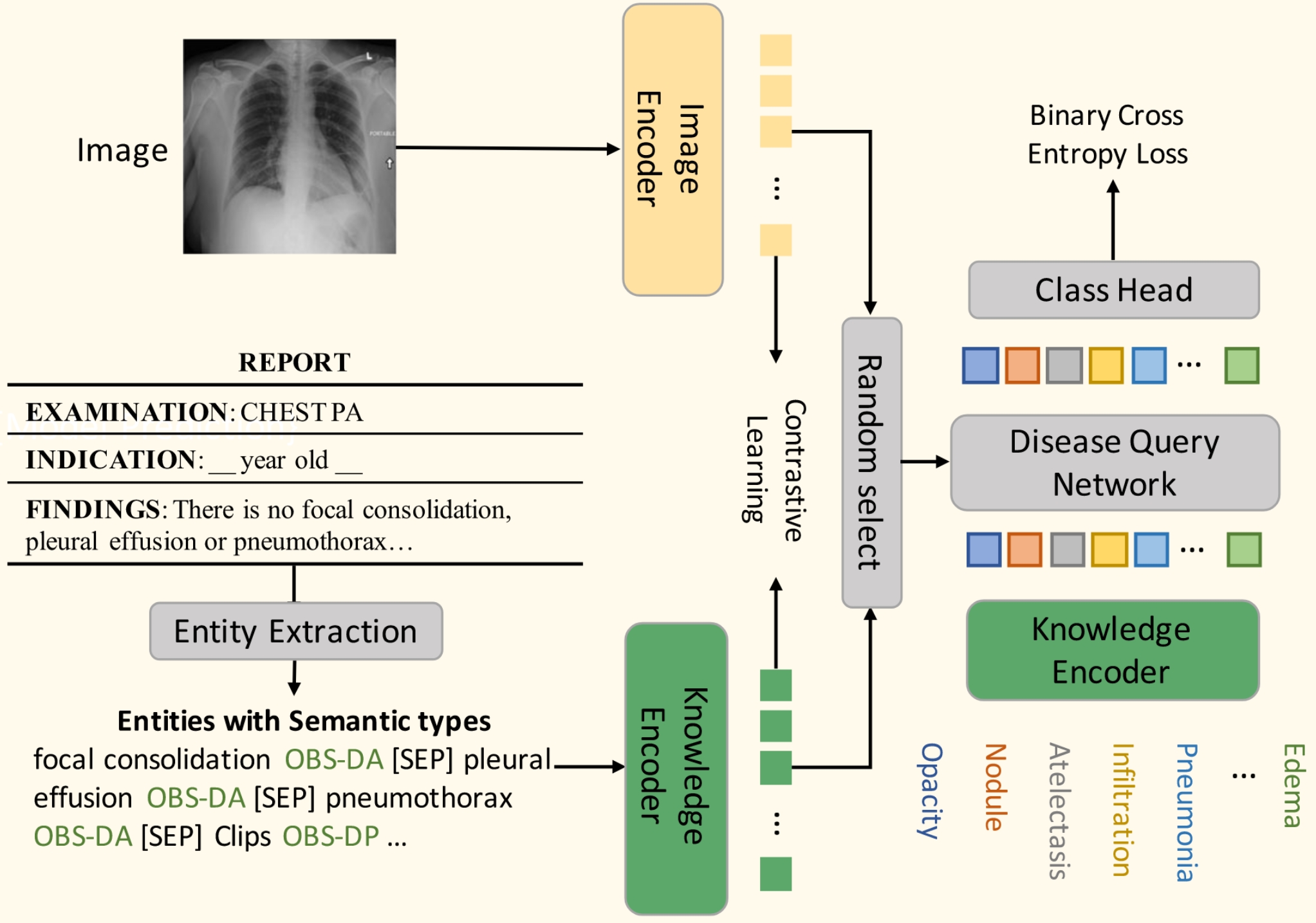}
    \subcaption{}
  \end{minipage}
  \begin{minipage}[t]{0.49\linewidth}
    \large
    \centering
    \includegraphics[width=\linewidth]{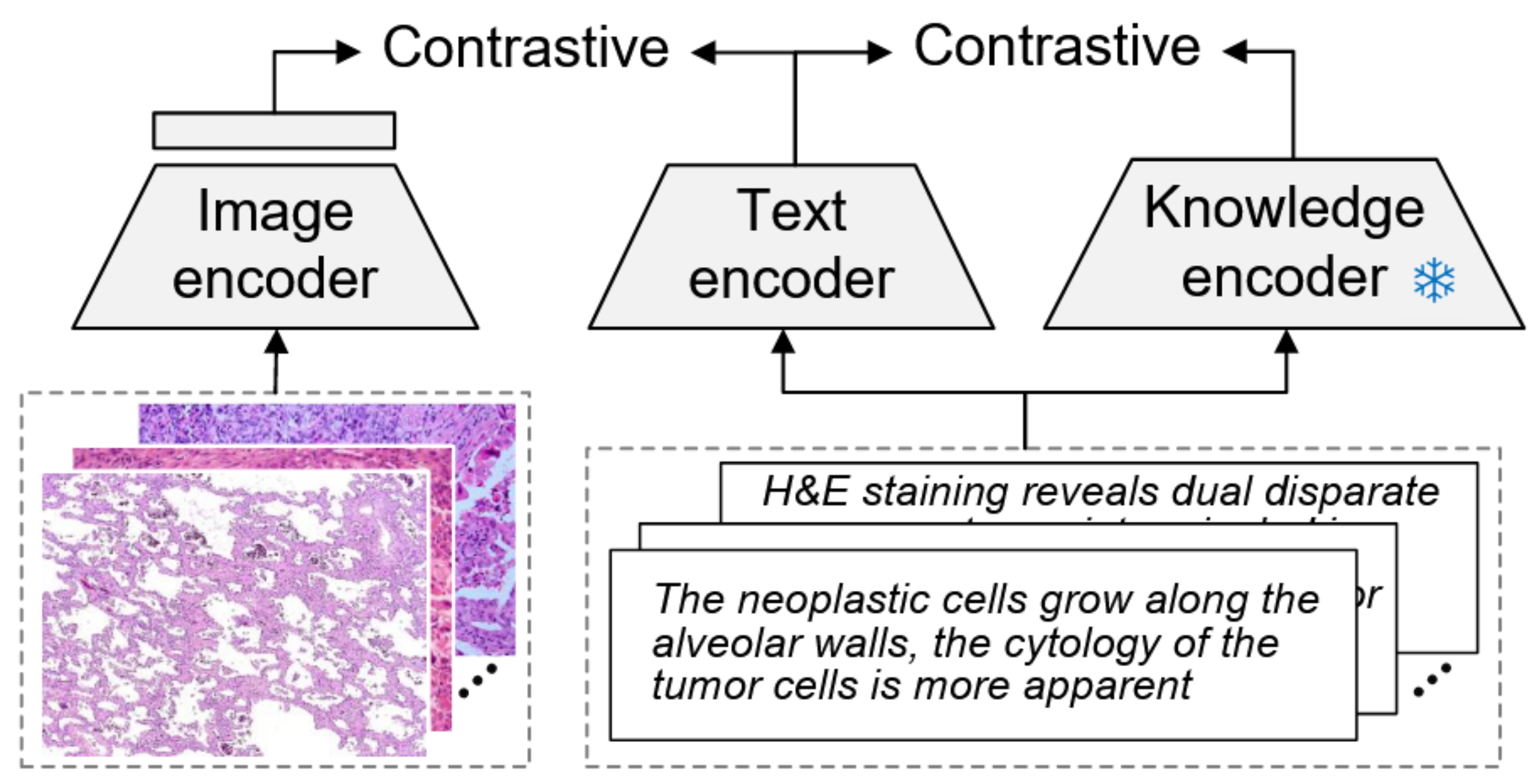}
    \subcaption{}
  \end{minipage}
  \caption{Illustration of CFM pre-training with structural knowledge fusion.
  (a) The second pre-training stage of KAD utilizes contrastive loss to align image and entity content representations and employs cross entropy for disease prediction. Taken from ~\cite{zhang2023knowledge}.
  (b) The second pre-training stage of KEP. Taken from ~\cite{zhou2024knowledge}.
  }
  \label{fig_knowledge}
\end{figure}

\myitem~\textbf{Structural Knowledge Fusion.}
To inject professional knowledge of the medical domain to enhance the semantic representations,
knowledge-fused pre-training is proposed.
They usually have two pre-training stages, where the first is to incorporate domain-structured knowledge to optimize the knowledge encoder and the second is for image-text alignment with the pre-trained knowledge encoder.
KAD~\cite{zhang2023knowledge} leverages medical knowledge to guide vision-language pre-training.
A well-established medical KG, \ie, UMLS, is introduced to fine-tune PubMedBERT by contrastive loss of concept-definition pairs and concept-relation-concept triplets.
Then the given raw X-ray reports are converted into contents of medical entities and their presence, using heuristically defined rules, RadGraph~\cite{DBLP:conf/nips/JainASTDBC0LNLR21}, or ChatGPT.
The pre-trained knowledge encoder is used to guide the visual representation learning by contrastive learning between representations of image and generated entity contents, effectively injecting the domain knowledge into the visual encoder.
Besides, the query disease is also input to incorporate randomly selected image or text entity content representations for disease prediction.
In the inference stage, by inputting unseen diseases, KAD can handle zero-shot disease prediction given an image.
KEP~\cite{zhou2024knowledge} curates a pathology \emph{knowledge tree} PathKT, consisting of three-level tree structures: tissue, disease, and attribute.
Each disease entity corresponds to several attributes, including disease synonyms, definitions, cytology and pathology features.
The knowledge encoder is updated by metric learning with AdaSP loss~\cite{zhou2023adaptive}, such that the representations of a specific disease and its attributes are close in the embedding space.
After that, the text encoder (initialized with the weights of the knowledge encoder) and image encoder are updated by image-text contrastive learning.
These methods of structural knowledge fusion are illustrated in Figure~\ref{fig_knowledge}.

\myitem~\textbf{Other Variants.}
For various aspects of multimodal medical applications, some other variants are proposed.
Previous methods could encounter many false negatives, meaning that images and reports from different patients could have the same semantics but are mistakenly treated as negative samples. So MedCLIP~\cite{DBLP:conf/emnlp/0008WA022} decouples image text pairs and conducts contrastive learning to reduce false negatives by introducing external medical knowledge.
To make full use of limited usable data and fix false negatives in contrastive learning, MedCLIP introduces UMLS to detect 14 main entity types for images with diagnosis labels. Multi-hot vectors of 14 dimensions from the extracted entities are obtained for images and texts, from which the semantic similarity is calculated.
The semantic similarity is viewed as \emph{soft targets} for model training rather than 0/1 labels in the original contrastive loss. In this way, unpaired data can also be taken into consideration.
To make the model capable of temporal information in the medical domain,
BioViL-T~\cite{bannur2023learning} exploits \emph{temporal correlations} by making prior images available for comparison to a given report.
The visual representations of the two images combine to make global and local contrastive learning, where an additional MLM is utilized for text-side pre-training.
PTUnifier~\cite{chen2023towards} introduces the \emph{soft prompts} to unify early-fusion and later-fusion medical vision-language pre-training,
making it compatible with different kinds of inputs, including image-only, text-only, and image-text pairs.
It constructs prompt pools for different modalities so that different inputs can select their corresponding prompts, improving the prompt diversity and the model scalability.
To consider the presence of community bias caused by different languages, Med-UniC~\cite{wan2024med} unifying \emph{cross-lingual} (English \& Spanish) medical multi-modal by diminishing dias.
Besides GCL for the vision-vision and vision-language alignment, cross-lingual text alignment regularization, including text augmentation, text-feature alignment, and text-to-text alignment, learns language-independent text representations and neutralizes the adverse effects of community bias on other modalities.

\subsection{Data Augmentation}
Medical texts are usually characterized by their specialized and condensed nature, making them difficult to understand by layman and neural models.
Therefore, several studies have introduced augmented text descriptions.
MedKLIP~\cite{DBLP:conf/iccv/WuZZWX23} focuses on the entities in the medical reports and adds entity descriptions.
The representations of these added descriptions are fused with image features to make predictions of entity existence and its location.
Based on it, MAVL~\cite{phan2024decomposing} further expands the description of disease entities to multiple visual aspects, including pattern, texture, opacity, border, location, shape, and fluid presence, where GPT-4 is utilized to programmatically generate descriptions of these aspects.
Beyond the loss functions of MedKLIP, it introduces another contrastive target to align visual representation and that of each entity aspect's description, empowering MAVL with the ability of zero-shot recognition of unseen diseases.
DeViDe~\cite{luo2024devide} utilizes publicly-available Mixtral-8×7B~\cite{jiang2024mixtral} to collect and process radiographic descriptions, for a specific entity or a disease.

Considering most augmentation techniques tend to narrow their focus, prioritizing either text or image augmentation, rather than blending the two.
PairAug~\cite{xie2024pairaug} designs a pairwise augmentation approach that contains an inter-patient augmentation (InterAug) branch and an intra-patient augmentation (IntraAug) branch.
Specifically, the InterAug branch generates radiology images using synthesised yet plausible reports derived from an LLM.
IntraAug branch uses newly generated reports to manipulate images.
This process facilitates the generation of new paired data for each individual with diverse medical conditions, where ChatGPT is used to report modification.

\subsection{Downstream Applications}

Based on pre-trained CFMs, many medical applications can be achieved, varying from uni-modal to cross-modal tasks.

\myitem~\textbf{Uni-modal Tasks.}
The jointly pre-trained image encoder and text encoder can be individually or jointly used for many uni-modal tasks, such as image classification, semantic segmentation, and object detection.
Medical image classification is usually to detect diseases in the image, based on GCL pre-training,
CFMs can do zero-shot image classification by calculating the similarity between image representation and text prompts with a specific disease, \eg, \emph{this is an image of \{disease\}} and \emph{\{disease\} presented in the image}~\cite{zhang2023biomedclip}.
Also, the image encoder can be frozen and the subsequent MLP is updated to fine-tune for image classification tasks, namely linear probing.
For semantic segmentation and object detection tasks, the pre-trained image encoder is initialized as the backbone encoder,
followed by a trainable task-specific decoder, like U-Net~\cite{ronneberger2015u} or ResUNet~\cite{diakogiannis2020resunet} and YOLOv3~\cite{redmon2018yolov3} for these two tasks, respectively.

\myitem~\textbf{Cross-modal Tasks.}
Pre-trained CFMs can also be used for many cross-modal tasks, including VQA, RG, ITR, TIR, and visual grounding.
As the CFMs generally have no ability for text generation, so its application for VQA mainly focuses on the classification setting, using VQA-RAD and SLAKE datasets~\cite{eslami2023pubmedclip,chen2023towards}.
The VQA model is usually initialized with a pre-trained CFM encoder and incorporates an MLP or specific decoder to make predictions.
For report generation, BioViL-T processes the prior report and both images (prior and current) with an encoder and an additional decoder is utilized for generation, where two broad categories, \ie, nearest-neighbour and auto-regressive can be used.
Retrieval-based tasks (ITR, TIR and disease retrieval) are directly realized by calculating similarities between the two modalities' representations.
Similarly, phase grounding calculates the similarity of representations between the text phase and the local image patch~\cite{boecking2022making,bannur2023learning}.

\begin{table*}[]
\setlength\tabcolsep{2pt}
\centering
\caption{Representative MLLMs in the general and medical domain. 
``/'' in the \emph{Datasets} \& \emph{Training Process} splits different pre-training stages.
Icons \smallsnow, \smallfire, and \smallminus\; denote the module is frozen, updating, and inexistence when training, respectively.
Their positions correspond image/adapter/language models.
RN is short for ResNet.
$\ddagger$ means to omit the pure LLM pre-training or the pure image encoder pre-training.
}
\label{tab_mllms}
\resizebox{\textwidth}{!}{
\begin{tabular}{rcclcl}
\toprule
\multirow{2}{*}{\textbf{Model}}&\multirow{2}{*}{\textbf{Time}}&\multirow{2}{*}{\textbf{Modality}}&\textbf{Image/Adapter/Language}&\multirow{2}{*}{\textbf{Size}}&\multirow{2}{*}{\textbf{Datasets (Training Process) \& Contribution}}\\
&&&\textbf{Model}&&\\
\midrule
Flamingo~\cite{alayrac2022flamingo}&04/2022&Natural&NFNet/QR+CA/Chinchilla&\scriptsize{3/9/80B}&M3W, ALIGN, LTIP, VTP (\snow\fire\snow); interleaved visual-textual data, few-shot in-context learning abilities.\\
CoCa~\cite{yu2022coca}&05/2022&Natural&ViT/CA/Transformer&\scriptsize{2.1B}&JFT-3B, ALIGN (\fire\fire\fire); CCL for image/text encoder, additional text decoder for language generation. \\
BLIP-2~\cite{DBLP:conf/icml/0008LSH23}&01/2023&Natural&ViT/QR/OPT, FlanT5&\scriptsize{3.1-12.1B}&COCO, CC3M, etc (\snow\fire\minus/\snow\fire\snow);  QFormer with cross-modal tasks, two-stage bootstrapping strategy. \\
LLaVA~\cite{DBLP:conf/nips/LiuLWL23a}&04/2023&Natural&ViT-L-14/LP/Vicuna&\scriptsize{13B}&CC3M/LLaVA-Instruct-158K, ScienceQA (\snow\fire\snow/\snow\fire\fire); instruction-following data using GPT-4.\\
MiniGPT-4~\cite{zhu2023minigpt}&04/2023&Natural&ViT-G-14/LP+QR/Vicuna&\scriptsize{13B}&Conceptual Caption, SBU, LAION, etc (\snow\fire\snow/\snow\fire\snow); only update a LP layer for alignment.\\

\midrule
SkinGPT-4~\cite{zhou2023skingpt}&04/2023&Camera&ViT-G-14/LP+QR/Vicuna&\scriptsize{13B}&SKINCON, Dermnet (\snow\fire\snow/\snow\fire\snow); based on MiniGPT-4, skin disease diagnoses by uploading photos.\\
PathAsst~\cite{sun2024pathasst}&05/2023&Pathology&ViT-B-16/LP/Vicuna&\scriptsize{$\sim$13B}& PathInstruct (\snow\fire\snow/\snow\fire\fire); specific model for pathology, capable of invoking eight sub-models.\\
MedBLIP~\cite{chen2023medblip}&05/2023&3D MRI&ViT-G-14/QR/BioMedLM &\scriptsize{$\sim$2.7B}&ADNI, NACC, etc (\fire\snow\fire/\snow\fire\fire); based on BLIP-2, LoRA, 2D vision encoder for 3D MRI scans.\\
LLM-CXR~\cite{lee2023llm}&05/2023&CXR&VQ-GAN (RN)/--/Dolly-v2-3B&\scriptsize{$\sim$3B}&MIMIC-CXR (\fire\minus\minus/\snow\minus\fire/\snow\minus\fire); no adapter, image tokens (input or output), CXR generation abilities. \\
BiomedGPT~\cite{zhang2023biomedgpt}&05/2023&Multiple&VQ-GAN (RN)/--/BART &\scriptsize{33-182M}&14 datasets (\fire\minus\fire/\snow\minus\fire); no adapter, unified FM, zero-shot transfer learning, diverse biomedical tasks. \\
XrayGPT~\cite{thawkar2023xraygpt}&06/2023&CXR&MedCLIP/LP/Vicuna&\scriptsize{13B}&MIMIC-CXR, OpenI (\snow\fire\snow/\snow\fire\snow); based on MiniGPT-4, answer open-ended questions about CXR.  \\
LLaVA-Med~\cite{li2024llava}&06/2023&Multiple&ViT-L-14/LP/Vicuna &\scriptsize{7/13B}&LLaVA-Med-Align/LLaVA-Med-Inst (\snow\fire\snow/\snow\fire\fire); based on LLaVA, instructions using GPT-4.\\
Med-Flamingo~\cite{moor2023med}&07/2023&Multiple&ViT-L-14/QR+CA/LLaMA&\scriptsize{8.3B}&MTB, PMC-OA (\snow\fire\snow); based Flamingo, few-shot in-context learning abilities, multi-image input ability. \\
Med-PaLM M~\cite{tu2024towards}&07/2023&Multiple&ViT/LP/PaLM&\scriptsize{12-62B}&MultiMedBench (\fire\fire\fire); generalist biomedical AI system, closed-source, can handle multiple modalities. \\
RadFM~\cite{wu2023towards}&08/2023&Radiology&3D ViT/QR/MedLLaMA&\scriptsize{14B}&MedMD/RedMD (\fire\fire\snow/\fire\fire\fire); 2D\&3D, tuning on very large-scale datasets, multi-image input ability. \\
RaDialog~\cite{pellegrini2023radialog}&10/2023&CXR&RN50/QR/Vicuna&\scriptsize{$\sim$7B}&MIMIC-CXR/image-grounded instruct data(\snow\fire\snow/\snow\snow\fire); LoRA, radiology RG \& interactive dialog.\\
Qilin-Med-VL~\cite{liu2023qilin}&10/2023&Multiple&ViT-L-14/LP/Chinese-LLaMA2 &\scriptsize{$\sim$13B}&ChiMed-VL-Align/ChiMed-VL-Inst (\snow\fire\snow/\snow\fire\fire); Chinese medical MLLM, multiple image modalities.\\
MAIRA-1~\cite{hyland2023maira}&11/2023&CXR&RAD-DINO/LP/Vicuna-7B&\scriptsize{$\sim$7B}&MIMIC-CXR (\snow\fire\fire); larger image resolution (518×518), leveraging GPT-3.5 for data augmentation. \\
PathChat~\cite{lu2023foundational}&12/2023&Pathology&ViT-L/LP+QR/LLaMA2&\scriptsize{$\sim$13B}&Alignment/instruction dataset (\snow\fire\snow/\snow\fire\fire); vision-language generalist assistant for human pathology. \\
MedXChat~\cite{yang2023medxchat}&12/2023&CXR&ViT-L-14/--/LLaMA&\scriptsize{$\sim$7B}&MIMIC-CXR (\snow\minus\fire); no adapter, LoRA, finetuned Stable Diffusion model for text-to-CXR synthesis.  \\
CheXagent$^{\ddagger}$~\cite{chen2024chexagent}&01/2024&CXR&EVA-CLIP-g/LP+QR/Mistral&\scriptsize{8B}&MIMIC-CXR, CheXinstruct, etc (\fire\fire\minus/\snow\fire\snow/\snow\fire\fire); multi-stage training, systematical evaluation. \\
CONCH~\cite{lu2024visual}&03/2024&Pathology&ViT-B-16/CA/Transformer &--&PubMed, Internal Data (\fire\fire\fire); based on CoCa, using diverse sources of histopathology images. \\
M3D-LaMed~\cite{bai2024m3d}&03/2024&3D CT&3D ViT/Pooling+LP/LLaMA2 &\scriptsize{6.9B}&M3D-Data (\snow\fire\fire); a versatile 3D MLLM, extra segmentation module for direct 3D image segmentation.  \\
Dia-LLaMA~\cite{chen2024dia}&03/2024&3D CT&ViT3D/QR/LLaMA2&\scriptsize{$\sim$7B}&CTRG-Chest-548K (\red{\snow\fire\fire}); LoRA, disease-aware attention, additional diagnostic information for RG. \\
LLaVA-Rad~\cite{zambrano2024training}&03/2024&CXR&BiomedCLIP/LP/Vicuna&\scriptsize{7B}&CXR-1M, MIMIC-CXR (\snow\fire\snow/\snow\fire\fire); LoRA, multi-stage tuning for lightweight MLLM. \\
WoLF~\cite{kang2024wolf}&03/2024&CXR&CLIP-ViT-L-14/?/Vicuna&\scriptsize{$\sim$7B}&MIMIC-IV, MIMIC-CXR (\snow\fire\fire/\snow\snow\fire); health-specific instruction and anatomy-specific knowledge. \\

\bottomrule
\end{tabular}
}
\end{table*}

\section{Multimodal LLMs (MLLMS)}
\label{sec_mllm}

Benefiting from the rapid development of LLMs, MLLMs, also known as visual language models (VLMs), have garnered significant attention from researchers owing to their powerful representational capabilities and remarkable proficiency in handling multimodal data~\cite{DBLP:journals/corr/abs-2306-13549}.
Their general modeling objectives are the next token prediction based on the image and previous text tokens:
\begin{equation}
\small
\mathcal{L}_{\rm MLLM} = \mathop{\mathbb{E}}_{i\in \mathcal{D},j\in \mathcal{M}} 
-\log p(\mathcal{T}_i^j|\mathcal{V}_i,\mathcal{T}_i^{<j}).
\end{equation}
From both theoretical and application standpoints, there exist distinct differences between CFMs and MLLMs:
1) CFMs are generally tuned based on the image-text pair data, whereas MLLMs focus on the multimodal instruction-following data;
2) GCL is the main objective for the CFMs while MLLMs are to generate text based on multimodal inputs;
3) CFMs are typically used for discriminative tasks, but MLLMs are more commonly used for generative tasks.
We summarize some typical general MLLMs and noted medical MLLMs in Table~\ref{tab_mllms}.

\subsection{Modality Encoder and Cross-modal Adapter}

MLLMs employ a straightforward yet effective method for building FMs.
This approach involves the use of an image encoder and language model, collectively known as modality encoders, which facilitates corresponding representations in latent spaces.
Additionally, a cross-modal adapter is introduced to align these representations of various modalities within a shared space.

\myitem~\textbf{Image Encoder.}
Pre-trained image encoders are typically employed to generate image representations, which are subsequently integrated with LLMs for multimodal tasks.
Commonly utilized pre-trained encoders include NFNet~\cite{brock2021high}, ViT~\cite{dosovitskiy2020image}, CLIP ViT~\cite{radford2021learning}, EVA-CLIP~\cite{fang2023eva} of the general domain.
Besides, to enhance the encapsulation of medical knowledge within the latent representations, pre-trained medical image encoders are employed in the creation of medical MLLMs.
For instance, the vision components of image-text pre-trained PathCap~\cite{sun2024pathasst}, BioMedCLIP~\cite{zhang2023biomedclip}, and BioViL-T~\cite{bannur2023learning} act as the image encoders of PathAsst~\cite{sun2024pathasst}, LLaVA-Med~\cite{li2024llava}, and RaDialog~\cite{pellegrini2023radialog}, respectively. Additionally, RAD-DINO~\cite{perez2024rad} employs CXR image pre-training, based on the self-supervised pre-training strategy of the DINO model~\cite{caron2021emerging}.

\myitem~\textbf{Language Model.}
Building on the powerful LLMs within the NLP domain, MLLMs employ them to process textual input and generate corresponding textual responses.
Both general-purpose and medical-specific MLLMs utilize the Transformer-style architecture as the text encoder.
This is evident in models such as OPT~\cite{zhang2022opt}, Flan-T5~\cite{chung2022scaling}, Vicuna~\cite{vicuna2023}, Mistral~\cite{jiang2023mistral}, LLaMA~\cite{touvron2023llama}, LLaMA2~\cite{touvron2023llama2}, and the Chinese-LLaMA2~\cite{cui2023efficient}, which is catered to the general domain.
In addition, there are several models tailored to medical languages, such as BioGPT~\cite{luo2022biogpt}, BioMedLM~\cite{venigalla2022biomedlm}, and MedLLaMA~\cite{wu2023pmc}, also in widespread use.

\begin{figure}[t]
\centering
\includegraphics[width=0.85\linewidth]{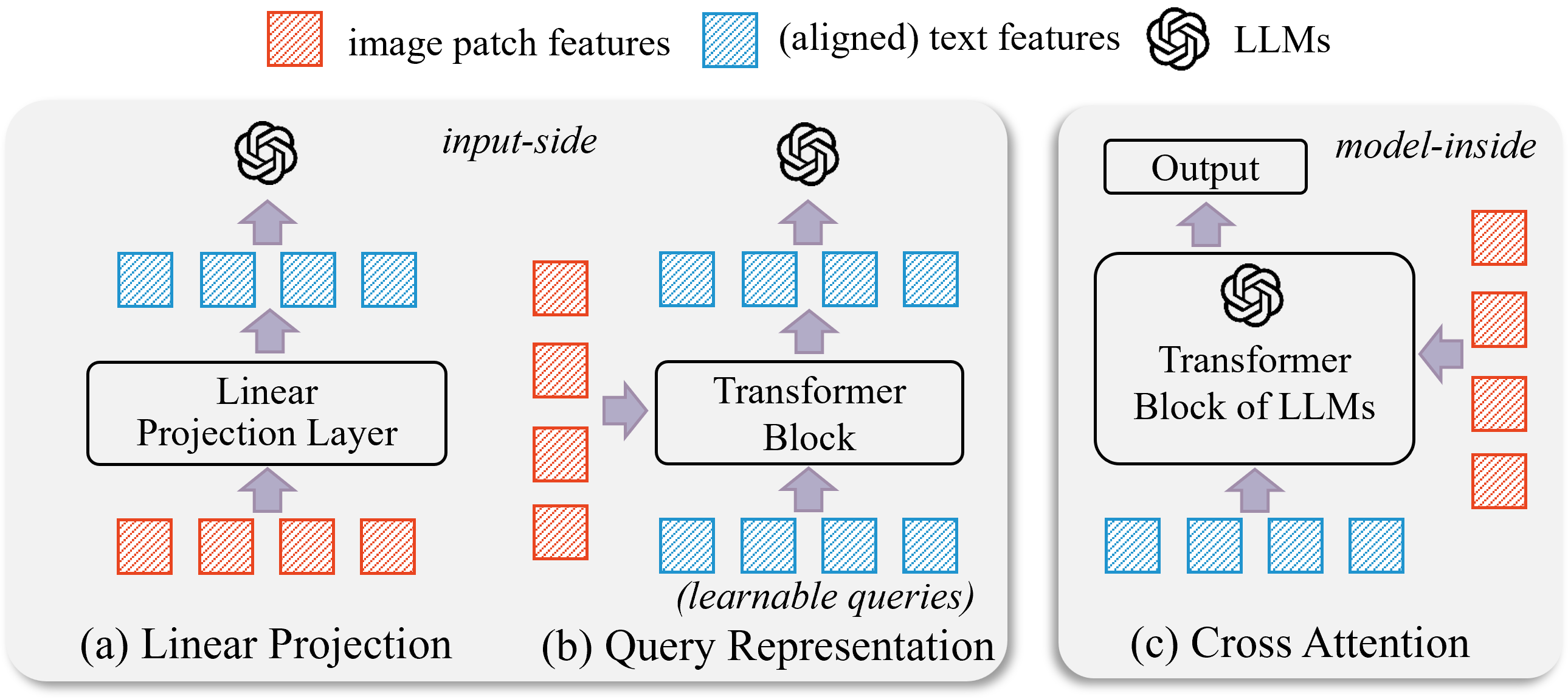}
\caption{Illustration of different types of adapters for MLLMs.
LR and QR are input-side while CA is inside of LLMs.}
\label{fig_adapter}
\end{figure}

\myitem~\textbf{Cross-modal Adapter.}
The cross-modal adapter serves as a connector between image and text representations within MLLMs.
Primarily, it encompasses three categories: \emph{Linear Projection}, \emph{Query Representation}, and \emph{Cross Attention},
as illustrated in Figure~\ref{fig_adapter}.
The first two types are handled during the input stages, transforming the hidden image representation into virtual token embeddings that align with text token embeddings.
The last type typically manifests within the internal computational procedures of the LLM~\cite{DBLP:journals/corr/abs-2306-13549}.
 
\textbf{Linear Projection (LP)}.
Typically, it employs a one or two-layer MLP to transform the image embedding space into a textual one.
Simple yet effective, LLaVA~\cite{DBLP:conf/nips/LiuLWL23a} first introduces it in the general MLLMs and has since been adopted by medical MLLMs,
such as PathAsst~\cite{sun2024pathasst}, XrayGPT~\cite{thawkar2023xraygpt}, LLaVA-Med~\cite{li2024llava}, and MAIRA-1~\cite{hyland2023maira}.

\textbf{Query Representation (QR)}.
This approach establishes learnable query representations for images,
which are then combined with textual token representations as inputs for LLMs.
The quantity of queries is usually significantly less than the number of image patches, thereby reducing computational efforts and enhancing efficiency.
The \emph{perceiver resampler} in Flamingo~\cite{alayrac2022flamingo} is designed to handle a flexible number of visual varying-size features (typically large) that are obtained from the vision encoder, generating a reduced number of visual outputs.
Similarly, QFormer, \ie, Querying Transformer, proposed in BLIP-2~\cite{DBLP:conf/icml/0008LSH23} is to unify the image information in the language space.
It first introduces several learnable queries that interact with image features using cross attention.
To align the multimodal input, three pre-training objectives are usually used before fine-tuning, \ie, image-text contrastive learning, image-grounded text generation, and image-text matching.
Regardless of the size of the visual encoder, the QFormer's final output length remains constant, \eg, 32 for BLIP-2 and MedBLIP~\cite{chen2023medblip}, significantly diminishing the computational load.

\textbf{Cross Attention (CA)}. Inspired by the self-attention of Transformer,
the cross attention approach assigns varying roles to image and language representations as \emph{query}, \emph{key}, or \emph{value} during the computational process of the LLM.
For example, the \emph{GATED XATTN-DENSE} in Flamingo~\cite{alayrac2022flamingo} views image representations as the \emph{key} and \emph{value}, the language counterpart as the \emph{query}.
Similarly, CoCa~\cite{yu2022coca} and CONCH~\cite{lu2024visual} employ cross attention for image information incorporation in the text decoder.

\subsection{Tuning Process \& Technical Details}

Based on the image encoder and language models, 
the implementation of MLLMs is achieved through the use of \emph{pre-training} and \emph{instruction-tuning} strategies~\cite{shrestha2023medical}.
The primary objective of pre-training is typically to adapt initializations from the general domain to the specialized medical one and to bridge distinct modalities.
Instruction refers to the task description and the goal of instruction tuning is to enhance a model's comprehension of user instructions and execute the tasks accordingly.
Through this process, MLLMs are capable of generalizing to previously unseen tasks with new instructions, thereby enhancing zero-shot performance~\cite{DBLP:journals/corr/abs-2306-13549}.
As the representative study, LLaVA-Med~\cite{li2024llava} is first fine-tuned on LLaVA-Med-Align (600K biomedical image-text pairs) to update linear transformation layer and then carries out the instruction-tuning on LLaVA-Med-Inst (60K image-text responses collected using GPT-4).
For parameter-efficient tuning, the LoRA~\cite{hu2021lora} technique, \ie, low-rank adaptation, could be utilized~\cite{chen2023medblip,pellegrini2023radialog,chen2024dia,zambrano2024training}.

\textbf{General MLLMs.}
Beyond the contrastive loss in CLIP, CoCa~\cite{yu2022coca} introduces the caption loss in the architecture.
The model comprises an image encoder, a text decoder, and a multimodal text decoder, capable of handling text generation tasks.
Flamingo~\cite{alayrac2022flamingo} is groundbreaking research in the realm of general MLLM,
which has been trained on vast multimodal web corpora containing arbitrarily \emph{interleaved} text and images.
This approach is vital for equipping Flamingo with in-context few-shot learning abilities.
In its framework, both the pre-trained vision encoder and language model are frozen.
Two novel components, namely the \emph{perceiver resampler} and \emph{GATED XATTN-DENSE}, are incorporated to effectively bridge the gap between powerful vision-only and language-only models.
The \emph{GATED XATTN-DENSE} block is integrated into every layer of the frozen LLM, facilitating cross attention between vision and language features.
BLIP-2~\cite{DBLP:conf/icml/0008LSH23} pre-trains a lightweight QFormer following a two-stage strategy to bridge the modality gap.
It first bootstraps vision-language representation learning from a frozen image encoder and
the second stage bootstraps vision-to-language generative learning from a frozen LLM.

LLaVA~\cite{DBLP:conf/nips/LiuLWL23a} and MiniGPT-4~\cite{zhu2023minigpt} are the pioneers of MLLMs tuning with image-text instruction data.
LLaVA is an initial endeavour to employ a language-only GPT-4 model for the creation of multimodal language-image instruction data known as LLaVA-Instruct-158K.
To effectively leverage the capabilities of both pre-trained LLM and visual model, a linear projection is introduced to align the vision features to the language counterpart.
LLaVA is pre-trained with only 1 epoch for feature alignment, where only parameters of the linear projection are optimized.
Then, keeping the visual encoder weights frozen, both pre-trained weights of the projection layer and LLM are updated on LLaVA-Instruct-158K for 3 epochs or on ScienceQA~\cite{lu2022learn}.
To realize numerous advanced multi-modal abilities demonstrated by GPT-4, MiniGPT-4 aligns a frozen visual encoder with the frozen Vicuna model using QFormer and a projection layer.

\begin{table*}[]
\setlength\tabcolsep{3pt}
\centering
\caption{Representative datasets for medical FMs, including for alignment and instruction-tuning.
Column \emph{LLM} indicates whether using LLM for dataset construction.
}
\label{tab_mllm_dataset}
\resizebox{\textwidth}{!}{
\begin{tabular}{rclclll}
\toprule
\textbf{Dataset}&\textbf{Modality}&\textbf{Scale}&\textbf{LLM}&\textbf{Remarks}&\textbf{Source}&\textbf{Time}\\
\midrule
PMC-15M~\cite{zhang2023biomedclip}&Multiple&15M&\XSolidBrush&collected from scientific papers, spanning thirty major biomedical image types.&PMC&03/2023 \\
OpenPath~\cite{huang2023visual}&Pathology&208K&\XSolidBrush&employ 32 pathology-related hashtags to retrieve relevant tweets.&Twitter, PathLAION&03/2023\\
PathCap~\cite{sun2024pathasst}&Pathology&207K&\Checkmark&pathology image-caption pairs, ChatGPT is employed to refine the captions. &PubMed, Books, Cytologists&05/2023\\
LLaVA-Med-Align~\cite{li2024llava}&Multiple&600K &\XSolidBrush&image-text pairs sampled from PMC-15M, images of multiple modalities for alignment.&PMC-15M&06/2023\\
MedMD~\cite{wu2023towards}&Radiology&16M&\XSolidBrush&multiple modalities of 2D\&3D, covering a wide range of anatomies with over 5000 diseases.&14 Sources&08/2023\\
ChiMed-VL-Align~\cite{liu2023qilin}&Multiple&580K &\Checkmark&covering X-ray, MRI, CT, radioisotope, mitotic, etc, translated by GPT-3.5 to Chinese.&PMC-OA, PMC-CaseReport&10/2023\\
CT-RATE~\cite{hamamci2024foundation}&3D CT&50K &\XSolidBrush&3D chest CT volumes and corresponding radiology text reports from hospital.&Internal Hospital&03/2024\\

\hline
PathInstruct~\cite{sun2024pathasst}&Pathology&180K&\Checkmark& description-based \& conversation-based (via ChatGPT), containing model-invoking part.&PathCap, Human Design& 05/2023\\
LLaVA-Med-Inst~\cite{li2024llava}&Multiple&130K&\Checkmark&using GPT-4 to generate multi-round Q\&A pairs, covering five modalities.&PMC-15M&06/2023\\
ChiMed-VL-Inst~\cite{liu2023qilin}&Multiple&469K&\Checkmark& QA pairs covering X-rays, CT scans, Echography, etc, translated by GPT-3.5 to Chinese.&PMC-VQA, PMC-CaseReport&10/2023\\
CheXinstruct~\cite{chen2024chexagent}&CXR&6.1M&\Checkmark&from existing or curating, five task categories, GPT-4 for translation and formulation.&65 Datasets&01/2024\\
M3D-Data~\cite{bai2024m3d}&3D CT&662K&\Checkmark&including tasks of VQA, vision language positioning, and segmentation, using Qwen-72B.&4 Datasets&03/2024\\
\bottomrule
\end{tabular}
}
\end{table*}

\textbf{Medical MLLMs.}
Motivated by the remarkable achievements of general MLLMs, medical MLLMs are developed with the aim of creating a highly efficient universal medical assistant.
For example, SkinGPT-4~\cite{zhou2023skingpt}, LLaVA-Med~\cite{li2024llava}, and Med-Flamingo~\cite{moor2023med} are the adaptions of MiniGPT-4, LLaVA, and Flamingo  within the medical field, respectively.
For human pathology,
PathAsst~\cite{sun2024pathasst} and PathChat~\cite{lu2023foundational} are developed through instruction tuning, utilizing pathology image-text data.
Taking PathAsst as an example, it constructs instruction-following data PathInstruct, which contains description-based and conversation-based, to revolutionize diagnostic and predictive analytics in pathology.
ChatGPT is utilized to generate conversational QA pairs based on image captions.
Special model-invoking instruction-following samples are also included.
Based on two-step tuning, PathAsst has the ability for VQA and conversation tasks related to pathology. Also, it can invoke sub-models for more comprehensive applications,
such as LBC (liquid-based cytology) classification and LBC detection.

Actually, a majority of medical MLLMs are predominantly focused on CXR or radiology modalities.
This concentration arises from the wealth of data available, exemplified by databases such as MIMIC-CXR.
A number of noteworthy studies exist in this direction, including LLM-CXR~\cite{lee2023llm}, XrayGPT~\cite{thawkar2023xraygpt}, MAIRA-1~\cite{hyland2023maira}, MedXChat~\cite{yang2023medxchat}, CheXagent~\cite{chen2024chexagent}, LLaVA-Rad~\cite{zambrano2024training}, and WoLF~\cite{kang2024wolf}.
These are constructed in line with the general strategies of MLLMs.
However, it's noted that LLM-CXR discards the design of the adapter.
It first maps the image to a fixed number of image tokens using pre-trained VQ-GAN~\cite{esser2021taming} based on the reconstructed L2 distance of the two features.
Then, these image tokens can be the input or output of LLMs, which is added to the token vocabulary and are random initialized.
In this way, LLM-CXR can not only handle the common generation task of CXR-VQA and CXR-to-report generation, but also the report-to-CXR generation, benefiting from VQ-GAN's ability to generate images. 

Beyond handling single modality, some studies are proposed to process multiple image modalities from the perspective of training data construction.
For example, BiomedGPT~\cite{zhang2023biomedgpt} covers pathology, dermatoscope, CT, radiology, and digital camera.
Med-PaLM M~\cite{tu2024towards} covers dermatology, mammography, radiograph, pathology, etc.
Beyond general 2D images, 3D image process models are also explored for richer spatial information modeling more scalable applications,
such as RadFM~\cite{wu2023towards}, M3D-LaMed~\cite{bai2024m3d}, and Dia-LLaMA~\cite{chen2024dia}.
Aiming to tackle a wide spectrum of clinical radiology tasks, RadFM is trained on large-scale comprehensive datasets MedMD and RadMD, covering various data modalities (X-ray, CT, MRI, etc), and tasks, featuring over 5000 diseases.
It possesses the capability to process both 2D and 3D images. For the 2D images, they are converted into 3D by merely extending an additional dimension. Subsequently, a 3D ViT is employed as the image encoder.
Similarly, M3D-LaMed and Dia-LLaMA also employ a 3D ViT as an image encoder.
During the tuning process of M3D-LaMed, the 3D image encoder remains frozen, while the proposed 3D spatial pooling perceiver and LLM with LoRA undergo updates.
Dia-LLaMA also introduces a disease prototype memory bank as a reference during diagnosis.
The disease-aware attention is proposed to extract disease-level representations from visual patches and disease-prototype contrastive loss is used to align these representations with learnable abnormal/normal prototypes.
Finally, the predicted diagnostic results can be converted into text prompts using a template description ``The \{disease name\} is [disease state]'' to view as an additional input to the LLMs to improve the diagnostic accuracy for infrequent abnormalities.

\subsection{Tuning Datasets}
\label{sec_tuningdata}

To realize medical MLLMs, there are usually two types of data utilized,
which can be summarized as alignment and instruction data.
They are concluded in the Table~\ref{tab_mllm_dataset}.

\begin{table*}[]
\setlength\tabcolsep{3pt}
\centering
\caption{Representative benchmark for MLLMs. Columns \emph{LLM} and \emph{H.} indicate using LLM and human experts for evaluation, respectively.
}
\label{tab_benchmark}
\resizebox{\textwidth}{!}{
\begin{tabular}{rclcclll}
\toprule
\textbf{Benchmark}&\textbf{Modality}&\textbf{Scale}&\textbf{LLM}&\textbf{H.}&\textbf{Tasks (Metrics)}&\textbf{Source}&\textbf{Time}\\
\midrule
MultiMedBench~\cite{tu2024towards}&Multiple&1M+&\XSolidBrush&\XSolidBrush&QA, RS, VQA, RG, IC (ACC, ROUGE-L, BLEU, F1-RadGraph, CIDEr-D, Macro-AUC, Macro-F1, etc).&12 Datesets&07/2023\\
RedBench~\cite{wu2023towards}&Radiology&137K&\XSolidBrush&\Checkmark&IC, RG, VQA, rationale diagnosis (ACC, BLEU, ROUGE, UMLS\_P, UMLS\_R, BERT-Sim, Human, etc).&13 Datasets&08/2023\\
PathQABench~\cite{lu2023foundational}&Pathnology&115&\XSolidBrush&\XSolidBrush&VQA (ACC, Pathologist evaluation for model comparison).&In-house Cases&12/2023\\
CheXbench~\cite{chen2024chexagent}&CXR&5K+&\Checkmark&\Checkmark&IC, VQA, RG, RS (ACC, ROUGE-L, CheXbert-S, BERT-S, RadGraph-S, GPT-4, Human).&7 Datasets&01/2024\\
OmniMedVQA~\cite{hu2024omnimedvqa}&Multiple&128K&\XSolidBrush&\XSolidBrush&VQA (ACC).&73 Datasets&02/2024\\
M3D-Bench~\cite{bai2024m3d}&3D CT&17K+&\Checkmark&\XSolidBrush&ITR, RG, VQA, positioning, and segmentation (Recall, Qwen-72B, ACC, BERT-Score, IOU, Dice, etc).&4 Datasets&03/2024\\

\bottomrule
\end{tabular}
}
\end{table*}

\textbf{Alignment Data.} It is used for pre-training to align image and text representations.
Medical image-text pairs, often found in textbooks or digital libraries, can be converted into alignment data through technical parsing and subsequent processing.
For example, PMC-15M~\cite{zhang2023biomedclip} is collected by PubMed Parser~\cite{achakulvisut2020pubmed} to process the XML files of PMC and extract captions and the corresponding figure references.
After that, items that lack figure references, exhibit syntax errors, or have missing information are systematically eliminated.
Upon it, LLaVA-Med-Align~\cite{li2024llava} is sampled from PMC-15M and ChiMed-VL-Align~\cite{liu2023qilin} is similar.

Given the vast number of medical images circulating online, particularly across social media platforms, it would be beneficial and promising to take them into account.
OpenPath~\cite{huang2023visual} collects de-identified pathology and their description on Twitter.
The 32 pathology-related hashtags are employed to retrieve relevant tweets, and strict protocols are followed regarding inappropriate sample removal and additional text cleaning.
Ultimately, 116K image–text pairs from Twitter posts and 59K pairs from the associated replies that received the highest number of likes are retained.
CT-RATE~\cite{hamamci2024foundation} comprises chest CT volumes and corresponding radiology text reports from a hospital. 
It includes about 50K reconstructed CT volumes from 25K distinct CT experiments conducted on 21K unique patients.

\begin{table}[]
    \centering
    \begin{tabular}{c}
                \begin{tcolorbox}[colback=gray!10,
                      colframe=black,
                      width=8.5cm,
                      boxrule=1pt,
                      arc=1mm, auto outer arc,
                     ]
                $\diamond$ \textbf{Interleaved data:} \blue{[Text1]}, \red{[Img1]}, \blue{[Text2]}, \red{[Img2]}, \blue{[Text3]}...\\ 
                $\diamond$ \textbf{Image-text pair:} \red{[Img]} -- \blue{[Description]}
                \end{tcolorbox} \\
                (a) Interleaved multimodal data and image-text pair. \\
                \begin{tcolorbox}[colback=gray!10,
                      colframe=black,
                      width=8.5cm,
                      boxrule=1pt,
                      arc=1mm, auto outer arc,
                     ]
                $\diamond$ \textbf{Coarse-grained Image Understanding:}\\
                Given an \red{[Img]}, the model is required to diagnose if the \blue{[Disease]} exists.\\
                $\diamond$ \textbf{Fine-grained Image Understanding:}\\
                Given the \red{[Img]}, localize the \blue{[Region]} of \blue{[Abnormality]}.\\
                $\diamond$ \textbf{Text Generation:}\\
                Given the \red{[Img]}, generate its \blue{[Caption]}.\\
                $\diamond$ \textbf{Question Answering:}\\
                Given the content of the given \red{[Img]}, answer the \blue{[Question]}.\\
                $\diamond$ \textbf{Miscellaneous:}\\
                Given the \red{[Img]}, select the text that best matches the image from \blue{[Options]}.
                
                \end{tcolorbox} \\
                (b) Instruction-following data, from CheXinstruct.\\
    \end{tabular}
    \caption{Multimodal data examples for medical FMs.}
    \label{tab_data}
\end{table}

\textbf{Instruction Data.} This type of data is used for the instruction tuning.
In the medical multimodal domain, it usually composes cross-modal tasks, such as VQA, RG, and coarse/fine-grained image understanding with text.
The creation of datasets typically involves using pre-defined prompt templates and answer text, feeding into LLMs to generate questions or to enhance instruction descriptions from a range of original data sources.
For example,
inspired by the LLaVA-Instruct~\cite{DBLP:conf/nips/LiuLWL23a} that process text parts of the multimodal data using LLMs, many studies adopt a similar manner for medical instruction data generation.
Biomedical instruction-tuning data LLaVA-Med-Inst~\cite{li2024llava} is collected using GPT-4, which is filtered from PMC-15M to retain the images that only contain a single plot.
Specifically, when presented with an image caption, instructions are formulated in a manner that encourages GPT-4 to produce multi-round questions and answers.
This is done in such a way that it appears as if GPT-4 can visualize the image itself, despite the fact that it only has access to the text.
CheXinstruct~\cite{chen2024chexagent} using about 28 public-available CXR datasets for generating instruction-tuning datasets.
It covers capability, task, dataset, and instance level.
It consists of five task categories according to their capabilities:
coarse-grained image understanding, fine-grained image understanding, question answering, text generation, and miscellaneous.

In summary, the current trend in datasets for MLLMs involves utilizing advanced LLMs to process text from various sources, such as medical textbooks, digital libraries, or original datasets.
The processed data is then transformed into corresponding outputs, ultimately creating image-text alignment data or image-text instruction data.
Their illustrations are shown as Table~\ref{tab_data}.

\subsection{Evaluation Benchmarks}

To comprehensively explore the capacities of medical MLLMs, several benchmark studies are proposed, such as MultiMedBench~\cite{tu2024towards}, RedBench~\cite{wu2023towards}, PathQABench~\cite{lu2023foundational}, CheXbench~\cite{chen2024chexagent}, and M3D-Bench~\cite{bai2024m3d}.
We give their details in Table~\ref{tab_benchmark}.
They usually integrate multiple datasets of the domain to perform a variety of tasks.
For evaluation metrics, three types are commonly utilized, including automatic statistical indicator (\eg, accuracy, ROUGE-L, BLEU, and CIDEr), AI evaluator (\eg, BERT-based and GPT-4), and human expert evaluator.

Typically,
MultiMedBench~\cite{tu2024towards} comprises more than 1 million data samples from 12 de-identified datasets of QA, RS (report summarization), VQA, RG, and IC, covering image modalities of pathology,
radiograph, genomics, mammography, etc.
RadBench~\cite{wu2023towards} encompasses five distinct tasks, including modality recognition, disease diagnosis, VQA, RG, and rationale diagnosis.
It has undergone meticulous manual verification to ensure data quality.
It also introduces two additional medical metrics for evaluation, \ie, UMLS\_P (precision) and UMLS\_R (recall), which aim to measure the overlapping ratio of medical-related words between ground truth and predicted response.
The medical-related words are extracted from them by using UMLS.
PathQABench~\cite{lu2023foundational} and OmniMedVQA~\cite{hu2024omnimedvqa} utilize the curated VQA datasets for comprehensive evaluation.
CheXbench~\cite{chen2024chexagent} has two evaluation axes, \ie, image perception and textual understanding.
The former utilizes six tasks across seven datasets, including view classification, binary disease classification, single disease identification, multi-disease identification, VQA, and image-text reasoning.
They are all in the format of multiple-choice and then the accuracy is viewed as the evaluation metric.
The latter evaluates the ability of models to generate and summarize text, where a combination of automated metrics (including GPT-4) and human expert evaluations (completeness, correctness, and conciseness) are utilized.

\subsection{Medical MLLM Applications}

In accordance with general MLLMs, medical MLLMs are typically employed to generate text responses from multimodal input, including medical visual chat, VQA, and RG.
For instance, Dia-LLaMA~\cite{chen2024dia} is capable of generating 3D CT reports.
LLaVa-Med~\cite{li2024llava} can handle both medical visual chat and VQA tasks, covering CXR, MRI, histology, gross, and CT domains.
The text outputs of MLLMs can also be enhanced by customizing inputs with domain knowledge.
To incorporate the expert prior knowledge of radiologists,
Kim \etal~\cite{kim2024enhancing} combined the original CXR image with its heatmap that highlights the precise focal points and duration of a radiologist’s attention, which provide extra human intelligence to MLLMs.
The experiment results demonstrate performance improvement.

Beyond only generating text responses, medical MLLMs can also give visual image responses, e.g., image synthesis or segmentation with a followed image decoder.
Borrowing the capacity of pre-trained VQ-GAN that uses auto-encoding architecture for CXR images, LLM-CXR~\cite{lee2023llm} could generate CXR images using LLM output of predicted virtual images codes by VQ-GAN decoder.
MedXChat~\cite{yang2023medxchat} can implement text-to-CXR synthesis, where the pre-trained stable diffusion (SD) model~\cite{rombach2022high} is used as the foundational framework for CXR generation, which is fine-tuning on MIMIC-CXR dataset using the zero-convolution strategy~\cite{zhang2023adding} for the adaptation from general domain to the medical one.
The generated prompts by the MLLM are then input into the SD model to generate CXR images.
RO-LMM~\cite{kim2023ro} introduces a 3D Residual U-Net~\cite{cciccek20163d} for image processing.
A multimodal alignment module is used to integrate comprehensive information from the image encoder and the pre-trained LLM. Subsequently, a 3D image decoder is employed to generate the segmentation mask to realize LMM-assisted breast cancer
treatment target segmentation.
M3D-LaMed~\cite{bai2024m3d} implements referring expression segmentation using a promptable segmentation module, where the last layer embedding of the \emph{[SEG]} token is extracted if it exists in the output.
After processing by an MLP, SegVol model~\cite{du2023segvol} is set as the promptable segmentation module to ultimately produce the segmentation mask.

\section{Discussions of Current Studies}
\label{sec_discussion}

In this section, we will discuss to answer our proposed question from the following five perspectives.

\textbf{(1). Can the existing multimodal data support advancing intelligent healthcare?}
Broadly speaking, existing datasets for multimodal healthcare suffer from the issues of diversity, volume, and simplistic construction approaches, which typically hinder the further development of technologies.
From \S\ref{sec_databases}, current RG datasets lack diversity and representation.
Medical reports cover a wide variety of conditions, diseases, and clinical scenarios.
However, some datasets may be collected in specific fields or healthcare institutions, resulting in samples that are too specific and lack sufficient diversity.
Besides, medical reports often rely on rich clinical context. However, existing datasets may not provide sufficient contextual information, such as patient history records and other examination results, which may limit the accuracy and completeness of generation models.

From \S\ref{sec_databases}, medical VQA datasets encounter the following limitations:
1) Imbalance of quality and quantity.
Most datasets are gathered automatically. Despite the significant advancements in the NLP field, 
achieving 100\% accuracy when generating samples remains a challenge.
For instance, when three human experts verify samples from the MIMIC-Diff-VQA~\cite{hu2023expert} dataset, the dataset demonstrates an average correctness rate of 97.4\% and a minimum correctness rate of 95\%.
While the VQA-RAD~\cite{lau2018dataset} and RadVisDial-Gold~\cite{kovaleva2020towards} datasets are of high quality due to manual collection, their sizes are relatively small, with 3,515 and 500 samples, respectively.
Creating a dataset that meets the high standards of both quality and quantity is difficult, given the significant demand for medical professionals.
2) Relative simpleness of the question.
Given that the majority of questions are generated automatically based on predefined rules or patterns, the questions in existing medical VQA datasets tend to be simplistic and lacking in variety.
For example, \emph{is the lesion associated with a mass effect}, \emph{what imaging method was used}, and \emph{what type is the opacity} in VQA-Med-2018~\cite{DBLP:conf/clef/HasanLFLML18}, VQA-Med-2019~\cite{ben2019vqa}, and MIMIC-Diff-VQA~\cite{hu2023expert}, respectively.
Although SLAKE~\cite{liu2021slake} incorporates a KG to answer questions that require external medical knowledge, its reasoning schema is in a one-hop direct manner.
For wider application prospects, there is a requirement for more comprehensive QA pairs.

From \S\ref{sec_tuningdata}, it is observed that the maximum data scale is 16M, which is much less than that in the general domain (\eg, ALIGN~\cite{DBLP:conf/icml/JiaYXCPPLSLD21} with 1.8B image-text pairs and LAION-5B~\cite{DBLP:conf/nips/SchuhmannBVGWCC22}).
For instruction data, the samples are generated using prompt engineering with LLMs or collected from diverse datasets directly, which could lead to simplistic and bias issues.
Moreover, there is insufficient emphasis on fine-grained data, which is crucial for medical image perception and implementation of the technologies in reality~\cite{wang-etal-2023-fine}.

\textbf{(2). Do task-oriented methods effectively address the targeted task?}
These methods have achieved certain success with the rapid development of AI technologies, but they also confront several significant challenges.
Taking RG as an example,
existing models have made notable progress in this field.
However, these models are still constrained by the data bias inherent in current datasets, which impairs their ability to detect and accurately describe subtle anomalies. On the MIMIC-ABN dataset~\cite{ni-etal-2020-learning}, which exclusively contains anomaly descriptions from the MIMIC-CXR dataset, the performance of these models on various metrics is significantly degraded~\cite{ni-etal-2020-learning, DBLP:conf/emnlp/HouCXLL23}. For instance, the BLEU-4 and ROUGE scores of RECAP~\cite{DBLP:conf/emnlp/HouCXLL23} on MIMIC-CXR are 12.5\% and 28.8\%, respectively, while on MIMIC-ABN, they drop to 8\% and 22.3\%, respectively. Furthermore, these models often overlook individual variability and the influence of patient-specific clinical context on diagnostic conclusions. This lack of consideration for personalized clinical information limits the overall effectiveness and accuracy of the generated reports.
Additionally, while many models perform well on natural language generation metrics such as ROUGE and BLEU, these metrics do not measure clinical accuracy. Consequently, there remains a significant gap between the performance of existing models and the requirements of clinical practice in terms of prediction accuracy, as indicated by metrics like F1-Chexbert.
In the field of image generation, existing techniques still suffer from the challenges of quality, accuracy, and interoperability.
They often struggle to accurately manipulate specific details, which is critical for medical image analysis~\cite{lyu2022conversion,meng2022novel,lee2023llm}.

\textbf{(3). How do FMs contribute to intelligent healthcare?}
Beyond task-oriented methods that usually undertake one specific task on one modality,
FMs contribute to unifying multiple tasks (\eg, IC, ITR, TIR, VQA, and RG) and modalities.
Their advantages stem from the utilization of large-scale parameters and training data.
Nonetheless, it also introduces challenges, such as complex deployment and diminished efficiency for training and inference~\cite{shrestha2023medical,azad2023foundational}.

\begin{figure}[t]
  \centering
  \begin{minipage}[t]{0.49\linewidth}
    \large
    \centering
    \includegraphics[width=0.99\linewidth]{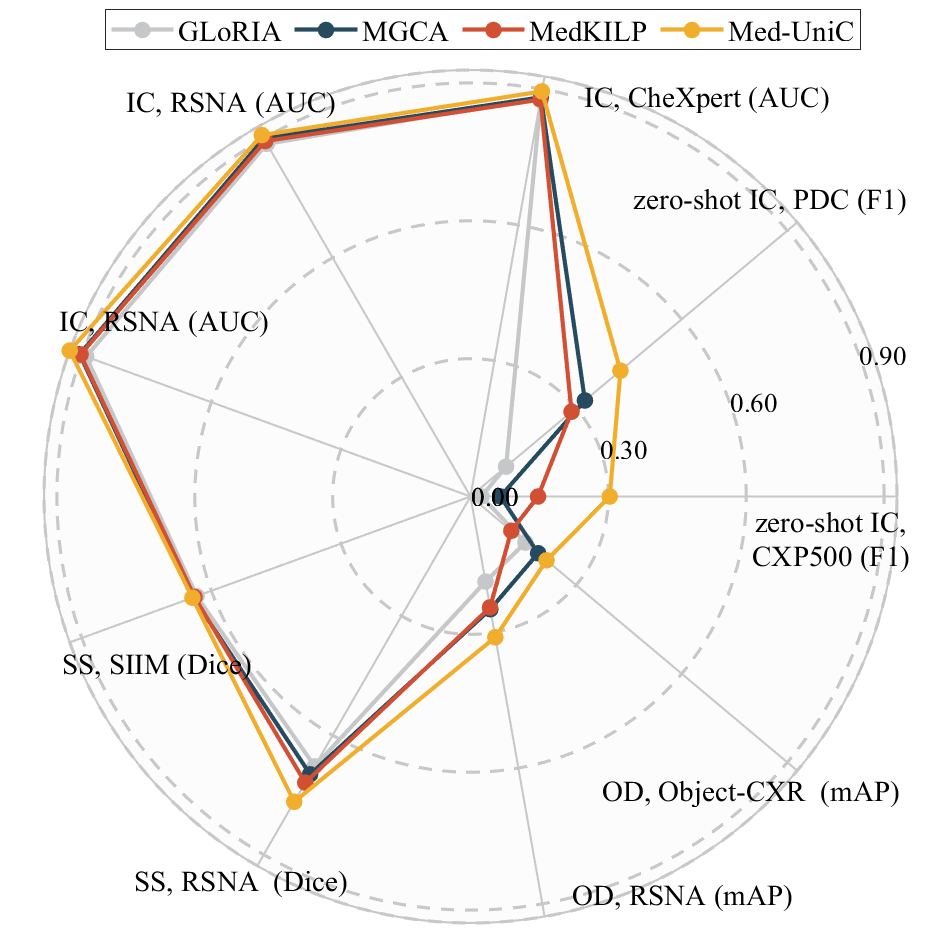}
    \subcaption{Results are obtained from Med-UniC~\cite{wan2024med}.}
    \label{fig1}
  \end{minipage}
  \begin{minipage}[t]{0.49\linewidth}
    \large
    \centering
    \includegraphics[width=0.99\linewidth]{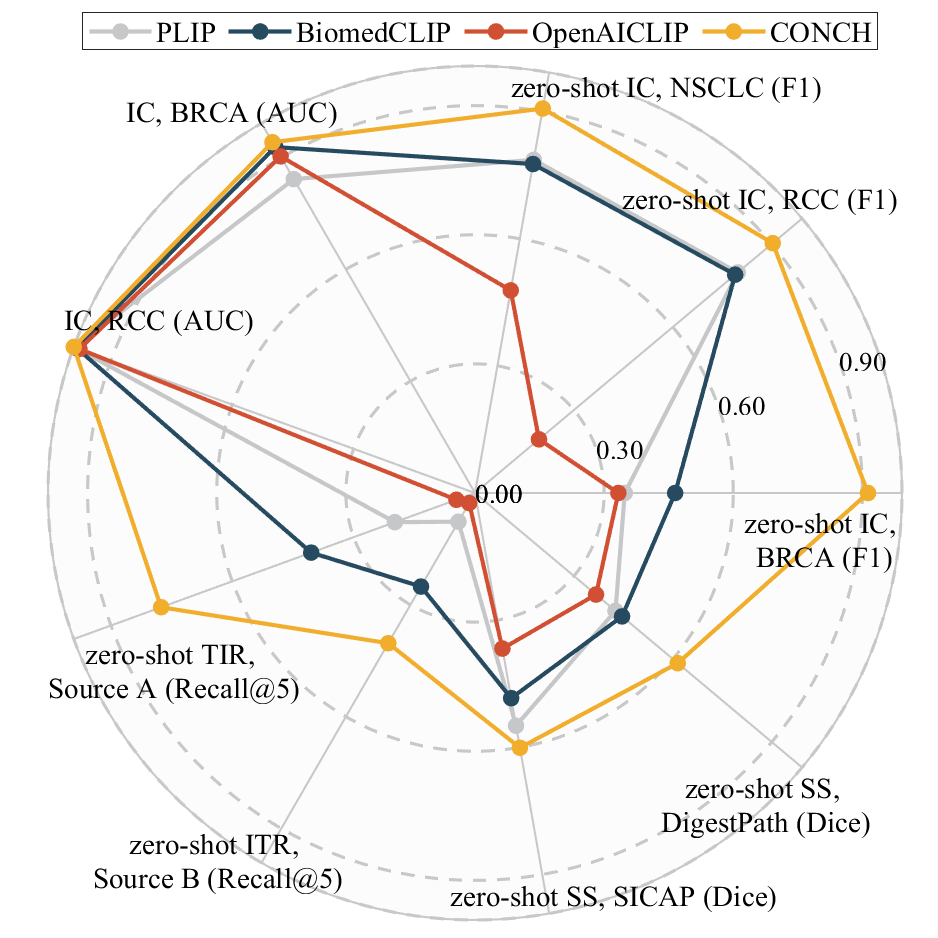}
    \subcaption{Results are obtained from CONCH~\cite{chen2024towards}.}
    \label{fig2}
  \end{minipage}
  \caption{Performance of medical FMs across various tasks.}
  \label{fig_cf_performance}
\end{figure}

We list some performance results of medical FMs in Figure~\ref{fig_cf_performance} and Table~\ref{tab_performance}.
It can be observed that the CFMs are consistently achieving improvement.
But they also need to be fine-tuned for better adapting downstream applications.
For instance, IC performance is usually better than the zero-shot IC.
Specifically, Med-UniC~\cite{wan2024med} only achieves about 30\% zero-shot IC F1 score on CXP500, and zero-shot TIR, ITR, and SS of CONCH~\cite{chen2024towards} is also relatively low, which would not suffice for practical applications.
We also explore the RG ability of MLLMs,
it can be seen that MAIRA-1~\cite{hyland2023maira} although achieves competitive results, it even doesn't surpass several non-MLLM models, especially for the clinical metrics that function as a transformation of text that preserves its semantics and prompts the model to concentrate on the report's critical elements without becoming excessively adapted to its style.
Beyond that, Hu~\etal~\cite{hu2024omnimedvqa} found that medical-specialized MLLMs even exhibit \emph{inferior} performance to those general-domain models, where BLIP-2~\cite{DBLP:conf/icml/0008LSH23} achieves the best performance for all tasks on average.
It indicates that the adaptation of general-purpose MLLMs using existing medical data does not result in emergent properties, underscoring the need for more adaptive and robust MLLMs within the medical domain.
In summary, though FMs bring new perspectives to healthcare and possess the ability to undertake multiple tasks, their ability to respond with high precision remains a challenge.

\begin{table}[t]
\centering
\caption{Findings generation performance on the MIMIC-CXR test set. Results are get from MAIRA-1. $\ddagger$ means the results are from non-MLLM models.}
\label{tab_performance}
\resizebox{0.5\textwidth}{!}{
\begin{tabular}{cccccc}
\toprule
\multicolumn{3}{c}{Lexical}&\multicolumn{3}{c}{Clinical}\\
\cmidrule(r){1-3} \cmidrule(r){4-6}
Metric&MAIRA-1&SOTA&Metric&MAIRA-1&SOTA\\

\midrule
ROUGE-L &\textbf{28.9} & 27.49& RadGraph-F1 &24.3 & \textbf{26.71} \\
BLEU-1 &\textbf{39.2}& 32.31 &RG\textsubscript{ER} &29.6& \textbf{34.7}$^\ddagger$\\
BLEU-4& \textbf{14.2}& 13.30$^\ddagger$ &CheXbert vector& 44.0 & \textbf{45.2}$^\ddagger$\\
METEOR &\textbf{33.3}& 16.8$^\ddagger$ &RadCliQ ($\downarrow$) &\textbf{3.10}& 3.277$^\ddagger$\\

\bottomrule
\end{tabular}
}
\end{table}

\textbf{(4). Do the current AI models present any ethical concerns?}
Despite promising advancements, the integration of AI into healthcare raises significant ethical and regulatory concerns. 
Issues like data privacy and model bias need careful attention to ensure AI systems are deployed responsibly in this sensitive area. 
Recent research indicates that FMs might allow data leakage, exposing personal health information when trained on specific data sources~\cite{huang2022large}. 
Additionally, biases in FMs often stem from uneven demographic distributions in the training data. 
As for bias, for example, the study shows that neural models trained on public chest X-ray datasets may underdiagnose in marginalized communities, such as female, black, Hispanic patients, and those insured by medicaid~\cite{seyyed2021underdiagnosis}. 
The broader ethical challenges with FMs, including fairness, accountability, and transparency, are complex and require ongoing attention to minimize ethical risks~\cite{he2023survey}.

Drawing from the superior capabilities of LLMs, medical MLLMs likewise adopt a vulnerability to hallucinations which manifest as irrelevant or factually incorrect responses for the input~\cite{chen2024detecting,pal2024gemini}.
This becomes a considerable concern in critical medical situations as there is little room for error.
Such errors or hallucinations can raise serious ethical issues for patients or doctors.
Different from the general domain, hallucinations in the medical domain can be characterized as multi-tasking, multi-faceted, and hierarchical, making them uniquely challenging and complex to address.

\textbf{(5). How do professionals assess current multimodal AI technologies?}
Despite the significant advancements in multimodal AI technologies within healthcare, medical professionals express concerns regarding potential challenges.
For example, 
Hu \etal~\cite{hu2024omnimedvqa} pointed out that despite medical MLLMs claim of robustness, they exhibit inferior performance to those in the general domain,
which reveals the limitations inherent in these medical models.
Therefore, to develop a more versatile professional, medical MLLMs should continuously incorporate specialized knowledge from various modalities, necessitating significant time and computational resources for refinement.
Acosta~\etal\cite{acosta2022multimodal} pointed out there existed challenges in curating higher-quality image–text datasets, the exceedingly high number of dimensions contained in multimodal health data, and multimodal model architectures.
They highlighted that despite the rapid progress in multimodal learning over recent years, present methods are \emph{unlikely} to be sufficient to address the major challenges.

Based on the above survey and discussions, we can answer the question \emph{has multimodal learning delivered universal intelligence in healthcare?}
The answer is \textbf{NOT}.
The existing research has several significant limitations (from data and technologies to performance and ethics) that need to be addressed to enhance its applicability in practical scenarios.

\section{Challenges and Future Directions}
\label{sec_challenge}

Drawing from the advancements in healthcare technology and the aforementioned discussions, we outline the following potential future directions.

\textbf{(1). High-quality \& Diverse Data.}
In reality, the current intelligent healthcare models are data-centric~\cite{zhang2024data}.
However, the datasets employed are somehow simplistic and homogeneous, as described in \S\ref{sec_discussion}(1).
The current success of general LLMs and MLLMs is based on the massive and diverse heterogeneous datasets~\cite{he2023survey,DBLP:journals/corr/abs-2306-13549}.
To benefit the healthcare community, high-quality and diverse data should be collected for more robust and flexible models applicable to reality.
The following aspects could potentially encompass it:
effective and varied integration of multiple multimodal datasets,
contextual data collection with backgrounds in real-life scenarios~\cite{bannur2023learning,hu2023expert},
user-oriented data construction with multiple image modalities,
and domain knowledge alignment data with fine-grained text or images.

\textbf{(2). Incorporating More Types of Modality.}
Present models mainly focus on medical images of radiology and pathology.
Approaches and techniques for modeling are comparatively well-established and offer the possibility of adaptation to other image modalities~\cite{ye2023continual}.
Medical audios (voice recording and stethoscope recording), videos (surgical and patient behavior videos), and time series (ECG, EEG, blood pressure, pulse, and other physiological signal data) can also be incorporated with others or with natural language (\eg, patients' condition descriptions or domain knowledge~\cite{zhang2023knowledge}) for more comprehensive modeling and more precise medical diagnosis and intervention.

\textbf{(3). Fine-grained \& High-resolution Image Modeling.}
The medical field necessitates precise visual modeling, given that typically only a minute fraction of the visual features bear relevance to the decision-making process, while the rest tend to be less informative.
Some fine-grained methods should be introduced to concentrate on the local representation of images, rather than only general representation.
The potential solution could be the application of multi-granularity, multi-scale, and hierarchical contrastive learning~\cite{lee2023llm,DBLP:journals/tkde/LinLZPHXZ23}.
Further, the model's capability to process high-resolution images is also crucial.
Present FMs mainly focus on low-resolution image processing, \eg, 224×224~\cite{tiu2022expert,li2024llava}.
Higher-resolution methods, such as BioViL~\cite{boecking2022making} (512×512) and MAIRA-1~\cite{hyland2023maira} (518×518), demonstrate performance improvements.
On one hand, high resolutions would benefit more detailed tissue structure information, clearer delineation of lesion boundaries, and precise quantitative assessment.
On the other hand, low-resolution vision encoders cannot directly handle ultra-high resolution medical images, \eg, high-resolution pathology whole-slide images (WSIs)~\cite{chen2024towards}.
Some advanced techniques in the general domain can be borrowed, employing high-definition encoders, or independently processing sub-images and subsequently integrating them,
which can be referred to Mini-Gemini~\cite{li2024mini} and Monkey~\cite{li2024monkey}.

\textbf{(4). Effective \& Efficient Knowledge Fusion.}
Vast amounts of domain knowledge in a graph-structured format are stored within meticulously curated knowledge bases, \eg, medical KGs (UMLS and PrimeKG~\cite{chandak2023building}).
Evidence confirms that distilling this particular knowledge into designated models is effective~\cite{zhang2023knowledge,zhou2024knowledge} and helps to relieve the hallucination of FMs~\cite{guan2024mitigating}.
However, the prevalent methods typically transform structured knowledge into token sequences and feed them into LMs for processing, which would cause information loss.
Seamlessly integrating this kind of knowledge into image or language models is challenging due to the inherent representation disparity between their original data.
A potential solution lies in learning general semantic tokens~\cite{liu2024rethinking} for discrete knowledge data, which can be directly incorporated into LMs.

\textbf{(5). Multimodal In \& Multimodal out.}
Current methods are limited to fixed-form inputs and outputs.
Task-oriented methods are usually inputted with one modality and output another modality or prediction.
CFMs usually need fine-tuning for downstream tasks and MLLMs concentrate on generating text responses about multimodal inputs.
Although Med-Flamingo~\cite{moor2023med} can be inputted with multiple images, its outputs are fixed to texts.
In reality, physicians and patients aspire to establish a diagnosis or predict health conditions based on historical records,
where the inputs and outputs would contain multiple medical images or other modalities (\eg, audio and video).
For instance, a patient diagnosed with scoliosis may wish to visualize spinal images after a period of treatment.
It can be referred to NExT-GPT (any-to-any MLLM)~\cite{wunext}, utilizing a variety of decoders for any combination outputs of text, image, and video.

\textbf{(6). Towards a Unified Model.}
Current models are often designed to process one or a few specific image modalities. However, for broader and more universal applications, it is crucial to develop unified medical models.
Given the vast range of medical modalities, organs, diseases, diagnoses, and medications,
employing a single-objective model for a specific task proves both challenging and economically inefficient in real-world applications.
Integrating primary and various image modalities, as well as both 2D and 3D images, is an urgent need and has great application prospects.
It can be realized from perspectives of data collection~\cite{zhang2023biomedgpt,wu2023towards} or model architecture~\cite{ye2023continual}.

\textbf{(7). Inspiring the Full Potential of FMs.}
Despite the proven effectiveness of FMs, researchers continue to explore their untapped potential. 
Early, they discovered that even small differences in similar natural language prompts could significantly impact model performance.
This led to various studies on different prompting methods to better utilize FMs, including automatically generated prompts, soft prompts, and pre-trained prompts~\cite{shin2020autoprompt}.
Subsequently, the Chain-of-Thought (CoT)~\cite{wei2022chain,zhang2023multimodal} method is introduced, where a model explicitly outlines the steps leading to its final answer, improving both transparency and performance. 
This method gives rise to comparable strategies, like Tree-of-Thought (ToT)~\cite{yao2024tree} and Graph-of-Thought (GoT)~\cite{besta2024graph}.
Additionally, retrieval augmentation-based methods have shown promise by incorporating external knowledge to boost FM performance.
Overall, these techniques can be tailored for medical FMs, unlocking their full capabilities and enhancing their practicality.

\textbf{(8). Comprehensive \& Unbiased Evaluation Protocol.}
Beyond the classification tasks that can be directly evaluated by accuracy,
current models, especially MLLMs, can generate open-ended answers with free text that is very hard to evaluate.
The demographic characteristics of the human evaluators, such as race, sex, age, and other factors, may introduce unconscious biases that affect their assessment of model performance~\cite{chen2024chexagent}.
Automatic statistical metrics like ROUGE-L and BLEU also fail to provide a satisfactory manner.
Comprehensive, standardized, and objective evaluation frameworks and metrics are urgent, yet they would be extremely laborious and resource-intensive.
Strong AI-based evaluation would provide a promising solution~\cite{bai2024m3d}.

\textbf{(9). Enhancing User-oriented Transparency \& Interpretability.}
As an area where there is a strong demand for accountability,
the models for medical applications are expected to be able to give explanations of the predictive outcomes that can convince doctors and patients.
Although current black-box models can achieve good prediction results, they are not transparent and therefore lack interpretability.
For this reason, some explainable reasoning frameworks~\cite{lin2023techs} can be referred to acquire internal logic rules for making predictions.
Additionally, with the power of MLLMs, questions can be decomposed into multiple human-understandable symbol subtasks\cite{xu2023symbol} and then solved individually to realize general interpretability.

\textbf{(10). Minimizing the Risks of Ethics.}
Some approaches need to be taken to minimize the risks of ethics, aiming to implement robust multimodal architectures~\cite{ma2024robust}.
Federated learning offers a solution by training models on local devices, preventing data exposure~\cite{li2020review}.
Methods like data resampling and model fine-tuning can help align these models with human values to mitigate bias issues~\cite{schulman2017proximal}.
As for the hallucination of MLLMs, it can be mitigated using technologies similar to the general domain, from data, model, training, and inference perspectives~\cite{bai2024hallucination}.

\section{Conclusion}
\label{sec_conclusion}

To explore the open research question \emph{has multimodal learning delivered universal intelligence in healthcare}, we carry out this study.
We first give a comprehensive review of the multimodal datasets, task-oriented techniques, and FMs.
Based on them, we carry out the discussion from five sub-questions.
Through our study, we discover that the current technologies fall short of realizing the goal of universal intelligent healthcare, highlighting the need for further exploration and development.
Additionally, we draw upon our findings to propose ten promising avenues for future research.
We anticipate that our study will stimulate extensive discussions among researchers regarding multimodal healthcare, thereby fostering the advancement of this field,
especially in the current age of swift advancement of LLMs technologies and their widespread applications.

{\small
\bibliographystyle{revision_ref}
\bibliography{reference}
}

\appendix

\section*{A. Dataset Details}

\subsection*{A.1 Datasets for Report Generation}
\label{sec_datasetrg}

There are several common datasets for report generation, as it is a very significant task where experts can communicate using language texts.
We summarize these studies in Table 2.
We can observe that they mainly focus on the radiology modality.
Current report generation datasets are mainly constructed using digital databases and libraries, \eg, PubMed.

At the beginning, IU X-ray~\cite{DBLP:journals/jamia/Demner-FushmanK16} undergoes an automated
de-identification process for X-ray
images and their responding reports, which is subsequently validated manually to ensure accuracy.
Each report consists of five sections:  \emph{indications}, \emph{findings},
\emph{impression}, \emph{manual encoding}, and \emph{MTI encoding}.
The former three denote the symptoms or reasons for examination, radiology observations, and final diagnosis, respectively.
The results of the manual and automatic coding of findings are embedded in the latter two.
ICLEF-Caption-2017~\cite{DBLP:conf/clef/EickhoffSHM17} collect 184.6K image-caption pairs from scholarly biomedical articles in PMC,
which is filtered from 3 million images automatically.
However, due to the fully automated nature of the construction method, the dataset contains a certain level of noise, notably 10$\sim$20\% of the images are either compound or non-clinical.
Based on ICLEF-Caption-2017, ICLEF-Caption-2018~\cite{DBLP:conf/clef/HerreraEAM18} further improves the quality and quantity of the dataset, which utilizes CNNs to decline noise and finally there are 232.3K image-caption pairs.
To explore the interaction between visual elements and semantic relationships inherent in radiology images,
Pelka~\etal\cite{DBLP:conf/miccai/PelkaKRNF18} automatically construct ROCO which is a fine-grained multimodal image dataset.
This intricately detailed multimodal dataset encompasses a variety of medical imaging modalities, such as Computer Tomography, Ultrasound, X-ray, Fluoroscopy, Positron Emission Tomography, Mammography, Magnetic Resonance Imaging, and Angiography.
It is filtered using pre-trained neural networks, and the final text is linked to UMLS CUIs and semantic types for data standardization and additional image interrelations.
PadChest~\cite{bustos2020padchest} is a large-scale and high-resolution CXR dataset.
For data gathering, 27\% of the reports are meticulously annotated by skilled physicians, while the rest is automatically labeled utilizing a supervised methodology that leverages the capabilities of an RNN with attention mechanisms.
Specifically, the reports are categorized using 174 distinct radiographic findings, 19 differential diagnoses, and 104 anatomical locations. These categories are structured as a hierarchical taxonomy and correspond to the standardized terminology of the UMLS.

For the pathology images,
PEIR Gross~\cite{DBLP:conf/acl/XingXJ18} collects images and their description captions from the Pathology Education Informational Resource (PEIR) digital library.
It only utilizes gross lesion images from 21 PEIR pathology sub-categories and each caption contains only one sentence that has an average of 12.0 words.
To enhance the dense supervision of tasks related to computational pathology, the ARCH dataset~\cite{gamper2021multiple} has been introduced.
It possesses a multiple instance feature, where a specific caption could be related to multiple images, referred to as a \emph{bag}.
The image-caption pairs are meticulously curated from PubMed articles and 10 different textbooks, followed by a comprehensive filtration process.

Given that medical figures in particular are typically complex and often comprised of multiple subfigures,
MedICaT~\cite{subramanian2020medicat} incorporates subcaptions and inline references to facilitate a comprehensive understanding of the relationship between text and figures,
which constructs a dataset of medical images in context.
Image inline references are typically located in the main body of the paper, significantly distanced from the pertinent figure, which often leads to them being overlooked. 
By employing MedICaT, research can be conducted to correlate each subfigure with its corresponding subcaption within a compound figure.
MedICaT surpasses previous datasets in its comprehensiveness, as it includes inline references for 74\% of the figures within the dataset.
Different from other main datasets that need to filter out the compound figures, ARCH and MedICaT are allowed compound figures.

To enhance the trustworthiness of the diagnostic methods as existing methods can only predict reports without accurate explanation,
FFA-IR~\cite{DBLP:conf/nips/LiCLWZWCLPLZLSV21} includes annotations of 46 categories of lesions with a total of 12,166 regions along with FFA images.
The annotation schema was developed by the ophthalmologists based on their expert knowledge, and covers most typical retinal lesions.
3D CT report dataset CTRG~\cite{tang2024work}, including brain and chest splits.
Subjective descriptions unrelated to the diagnosis are filtered out of the report.
The report is then split into a template and abnormal descriptions to facilitate model training.

\subsection*{A.2 Datasets for Medical VQA}
\label{sec_datasetvqa}

As medical images are typically accompanied by corresponding captions in the database, creating medical VQA datasets primarily involves selecting images, choosing sentences from the captions as answers, and finally generating questions.
We summarize the main characteristics of the current common-used dataset for medical VQA in Table 3.

As a groundbreaking study, VQA-Med-2018~\cite{DBLP:conf/clef/HasanLFLML18} introduces a semi-automatic approach to generate question-answer pairs from captions of the radiology images sourced from PubMed, released on the ImageCLEF platform\footnote{https://www.imageclef.org/} (based on ICLEF-Caption-2017).
A rule-based system is employed to generate possible question-answer pairs from captions and then the question ranking process is conducted to select the most appropriate question.
Finally, two expert human annotators carefully review all the generated samples to filter out noisy and incorrect ones.
The text similarity metrics BLEU as well as word-based and concept-based semantic similarity are introduced for evaluation.
Similarly to VQA-Med-2018, VQA-Med-2019~\cite{ben2019vqa}, VQA-Med-2020~\cite{DBLP:conf/clef/AbachaDHDM20}, and VQA-Med-2021~\cite{ben2021overview} automatically generate radiology VQA samples based on MedPix database.
They all generate specific questions about sentences in image captions using predefined rules or patterns.
It is noted that VQA-Med-2019 also utilize the accuracy metric of exact matching for generation evaluation, emphasizing the precise match between a predicted answer and the ground truth answer.
To guarantee the quality and investigate the question in a realistic scene,
VQA-RAD~\cite{lau2018dataset} dataset is carried out, which is the first manually constructed dataset.
Clinicians are naturally presented with questions regarding radiology images (CT, MRI, X-ray) and are expected to provide corresponding answers.

Different from the above generation datasets, the answer format in RadVisDial-Silver and RadVisDial-Gold~\cite{kovaleva2020towards} is structured as a multiple-choice selection among four options.
It indicates that they involve classification tasks and the question contents are from 13 abnormalities.
RadVisDial-Silver and RadVisDial-Gold are gathered through automated and manual means, respectively.
PathVQA~\cite{he2020pathvqa} is the first attempt to build a VQA dataset about pathology images.
It extracts pathology images and their captions from both electronic pathology textbooks and the Pathology Education Informational Resource (PEIR) Digital Library\footnote{http://peir.path.uab.edu/library/index.php?/category/2}.
Subsequently, sentence simplification, question transducer, and post-processing are introduced to generate questions.
SLAKE~\cite{liu2021slake}, a bilingual dataset that contains samples in both English and Chinese, emphasizes the semantic labels and a structured medical knowledge graph in its features.
It offers two types of semantic visual annotations for each radiology image: masks for semantic segmentation and bounding boxes for object detection.
In addition to basic clinical questions, it introduces compositional questions that demand multiple reasoning steps utilizing a personalized medical KG derived from OwnThink\footnote{https://www.ownthink.com}.
To illustrate the difference between the two medical images for monitoring changes in patients' physical conditions,
the MIMIC-Diff-VQA~\cite{hu2023expert} dataset is introduced.
The initial step involves gathering keywords, such as names of abnormalities and important attributes like location, level, and type, from the MIMIC-CXR dataset.
Next, an intermediate KeyInfo database is created, and study pair questions are generated based on specific patterns.
Each pair of images consists of a main image and a reference image, derived from distinct studies of the identical patient.
The selection of main and reference visits adheres strictly to the initial visit being designated as the \emph{reference} and the subsequent visit as the \emph{main} image.
To facilitate an all-inclusive assessment of large vision-language models within the medical domain, 
Hu~\etal~\cite{hu2024omnimedvqa} construct OmniMedVQA dataset.
The initial phase of this process involves the collection of 75 distinct medical classification datasets encompassing 12 fine-grained imaging modalities,
covering over 20 human anatomical regions.
Subsequently, unique QA templates are devised for each dataset to generate questions based on the original category labels.
The final step refines the QA pairs using GPT-3.5 and human validation checks.

\ifCLASSOPTIONcaptionsoff
  \newpage
\fi

\vfill


\end{document}